\pdfoutput=1

\documentclass[11pt]{article}

\usepackage{arXiv}

\usepackage[T1]{fontenc}
\usepackage{microtype}
\usepackage{times}
\usepackage{textcomp}
\usepackage{inconsolata}

\usepackage{arabtex}
\usepackage{utf8}
\setcode{utf8}
\makeatletter
\protected\def\begin#1{%
  \UseHook{env/#1/before}%
  \@ifundefined{#1}%
    {\def\reserved@a{%
       \@latex@error{Environment #1 undefined}\@eha
     }}%
    {\def\reserved@a{%
       \def\@currenvir{#1}%
       \edef\@currenvline{\on@line}%
       \@execute@begin@hook{#1}%
       \csname #1\endcsname
     }}%
  \@ignorefalse
  \begingroup
  \let\end\a@l@end
  \@endpefalse
  \reserved@a
}
\makeatother
\newcommand{\armini}[1]{%
  {\scriptsize\setcode{utf8}\<#1>}%
}

\usepackage{amsmath}
\usepackage{amssymb}
\usepackage{latexsym}

\usepackage{graphicx}
\usepackage{xcolor}
\usepackage{colortbl}
\usepackage{booktabs}
\usepackage{multirow}
\usepackage{array}
\usepackage{tabularx}
\usepackage{makecell}
\usepackage{subcaption}
\usepackage{adjustbox}

\usepackage{setspace}
\usepackage{fancyhdr}
\usepackage{tipa}
\usepackage{siunitx}
\usepackage{soul}
\usepackage{pifont}
\usepackage{xspace}
\usepackage{comment}

\usepackage{cuted}
\usepackage{caption}

\usepackage{tcolorbox}

\definecolor{rankone}{RGB}{180,230,180}
\definecolor{ranktwo}{RGB}{180,210,240}
\definecolor{rankthree}{RGB}{255,220,180}

\definecolor{ClassicalGreen}{HTML}{5FAD41}
\definecolor{LightGreen}{HTML}{C8E6B0}

\definecolor{MajorRed}{HTML}{C0392B}
\definecolor{LightRed}{HTML}{F5B8BA}

\definecolor{MinorOrange}{HTML}{E67E22}
\definecolor{LightOrange}{HTML}{FAD7A0}

\definecolor{RegPurple}{HTML}{7D6BB0}
\definecolor{LightPurple}{HTML}{D7BDE2}

\definecolor{HeaderNavy}{HTML}{1B2A4A}

\definecolor{MSABlue}{HTML}{D6EAF8}
\definecolor{MSADark}{HTML}{1A5276}

\definecolor{BestCountry}{RGB}{0,70,140}
\definecolor{BestModel}{RGB}{0,120,60}

\definecolor{paleAgreementBg}{RGB}{255,242,242}
\definecolor{frameBorderColor}{RGB}{220,200,200}

\newcommand{\cmark}{\ding{51}}
\newcommand{\xmark}{\ding{55}}

\newcommand{\Cmark}{\textcolor{blue}{\cmark}}
\newcommand{\Xmark}{\textcolor{red}{\xmark}}

\newcommand{\ourdataset}{\textsc{Bulbul}\xspace}

\newcolumntype{C}[1]{%
  >{\centering\arraybackslash}m{#1}%
}

\newcolumntype{E}[1]{%
  >{\raggedleft\arraybackslash}m{#1}%
}

\newcolumntype{H}{%
  >{\setbox0=\hbox\bgroup}c<{\egroup}@{}%
}

\newtcolorbox{agreementbox}{
  colback=paleAgreementBg,
  colframe=frameBorderColor,
  arc=3mm,
  boxrule=0.4mm,
  left=10pt,
  right=10pt,
  top=6pt,
  bottom=6pt,
  width=\textwidth,
  fonttitle=\bfseries
}

\title{%
  \raisebox{-2.1ex}{%
    \protect\includegraphics[
      height=4\fontcharht\font`\B
    ]{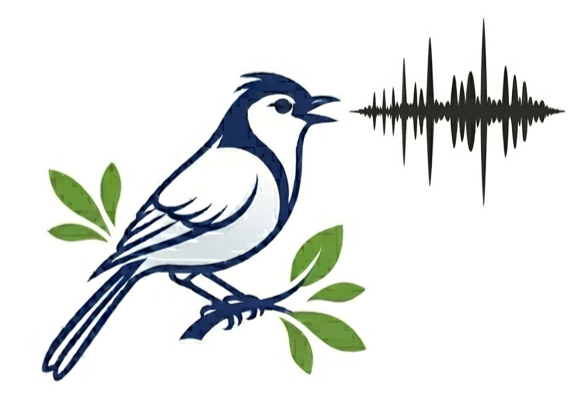}%
  }%
  \quad
  Bulbul: A Dataset for Dialectal Arabic Speech Recognition
}

\author{%
Ahmed Ashraf$^{1}$,
Aisha Alansari$^{1}$,
Fadel Al Abbas$^{1}$,
Nada Almarwani$^{2}$,
Samah Aloufi$^{2}$,
Saad Ezzini$^{1}$ \\
Maged S. Al-Shaibani$^{1}$,
Doaa Dalaq$^{1}$,
AbdelRahim A. Elmadany$^{3}$,
Muhammad Abdul-Mageed$^{3}$ \\
Mohamed Mehdi Trigui$^{1}$,
Dania Refai$^{1}$,
Layan Refai$^{4}$,
Mohamed Akrout$^{1}$,
Mustafa Jarrar$^{5,6}$ \\
Wasfi G. Al-Khatib$^{1}$,
Alaa Dalaq$^{1}$,
Darin El-Nakla$^{1}$,
Samir Abdaljalil$^{7}$,
Abdulrahman Al-Fakih$^{1}$ \\
Nour El Imane Zeghib$^{1}$,
Moussa REDAH$^{1}$,
Salmane Chafik$^{8}$,
Mohamed El-Attar$^{9}$,
Rima Grati$^{9}$ \\
Sarah Kohail$^{9}$,
Malak Alkhorasani$^{10}$,
Khadijah Al Safwan$^{1}$,
Ismail M. Mudhaffar$^{1}$,
Ali Altam$^{11}$ \\
Ahmed Al-Shaikh$^{1}$,
Adnan Saeed$^{12}$,
Hamzah Luqman$^{1}$ \\[12pt]
$^{1}$KFUPM,
$^{2}$Taibah University,
$^{3}$UBC,
$^{4}$PSUT,
$^{5}$Birzeit University,
$^{6}$HBKU \\[-0.1em]
$^{7}$Texas A\&M University,
$^{8}$Mohammed VI Polytechnic University,
$^{9}$Zayed University,
$^{10}$IAU \\[-0.1em]
$^{11}$Symbiosis International University,
$^{12}$Taiz University \\
\normalfont\small
\texttt{\{g202411740,aisha.ansari,hluqman\}@kfupm.edu.sa}%
}

\begin{document}

\maketitle


\begin{strip}
\centering

\includegraphics[
  width=\linewidth
]{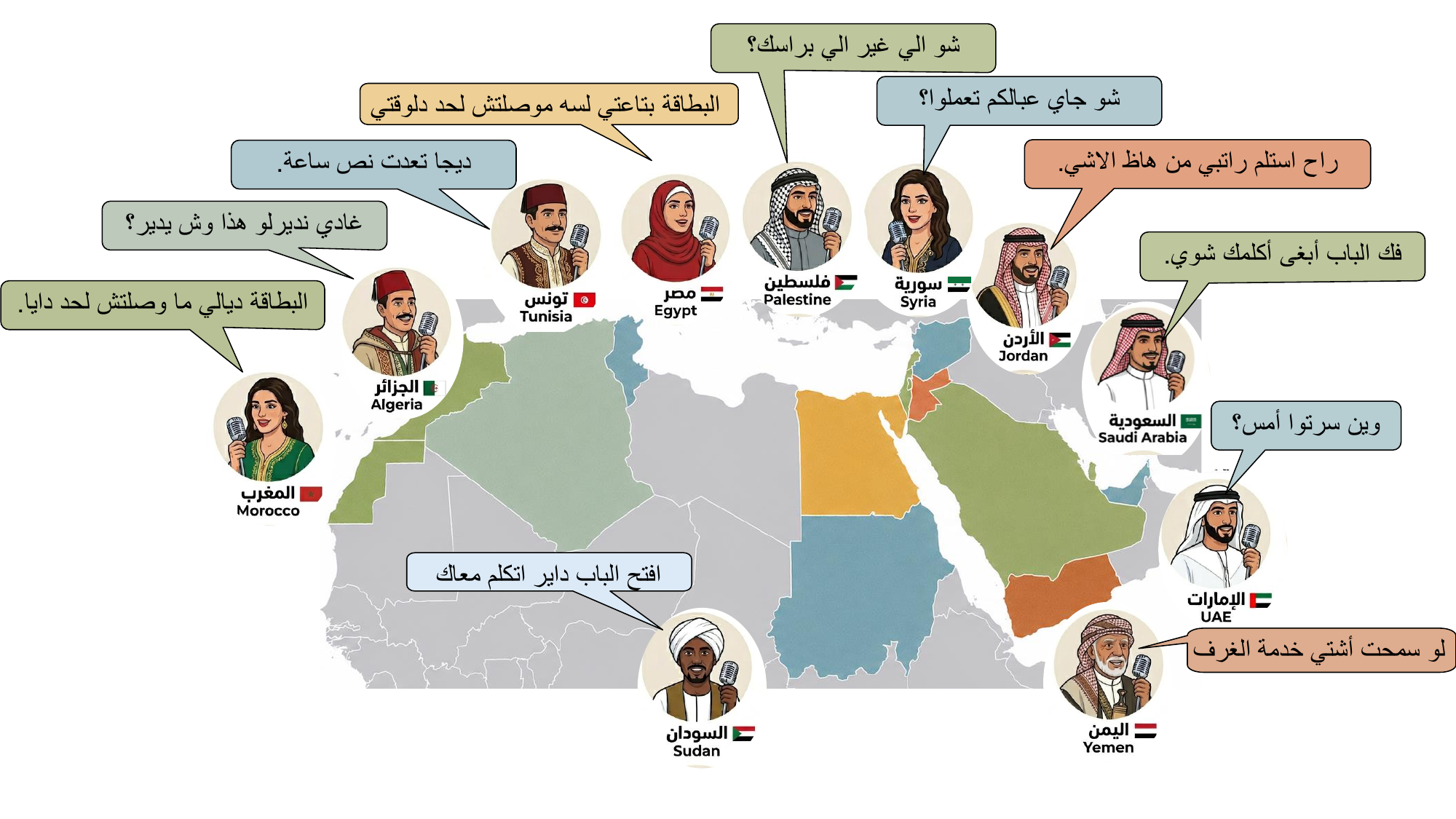}

\captionof{figure}{
  Overview of \ourdataset, covering 11 Arab countries and
  presenting representative examples from each dialect
  (English translations provided in
  Appendix~\ref{translation}).
}
\label{fig:abstract}

\end{strip}


\section*{Abstract}

Arabic automatic speech recognition (ASR) faces unique challenges due to diglossia, extensive regional dialect variation, and limited speech resources. Existing speech datasets often focus on single dialects or large-scale broadcast/web data, leading to trade-offs between linguistic diversity and annotation quality. We present \ourdataset, a multi-dialect Arabic ASR dataset collected from 275 speakers in 11 Arab countries. \ourdataset includes structured dialect and sub-dialect coverage, as well as recordings of classical Arabic and modern standard Arabic spoken by participants in their native dialectal accents to support accent-aware modeling. The quality of the recordings was ensured through a two-level human verification process. We further benchmark a range of recent ASR systems, establishing strong baselines for modern dialectal and accented Arabic ASR.

\section{Introduction} 
\label{sec_intrdocution}

Recent advances in automatic speech recognition~(ASR) have revolutionized spoken language technologies, demonstrating remarkable capabilities across diverse languages and domains. In Arabic, ASR systems must navigate a uniquely complex linguistic landscape characterized by immense phonetic and lexical diversity in dialects and subdialects \cite{alqadasi2025arabic,alshargi2019morphologically}. This variety results in limited and unbalanced speech resources, heavily skewed toward a few dialects, leading to poor generalization for underrepresented dialects \cite{djanibekov2025dialectal}.

\begin{table*}[t!]
\centering
\scriptsize
\setlength{\tabcolsep}{3.5pt}
\renewcommand{\arraystretch}{1.1}
\caption{Comparison of the \ourdataset~dataset with existing Arabic ASR datasets.}
\label{tab:dataset_comparison_new}

\begin{tabular*}{\textwidth}{@{\extracolsep{\fill}} >{\raggedright\arraybackslash}p{3.9cm} *{11}{c} @{}}
\toprule
Dataset & Hrs & Spk & Dial & Micro & Text Src & Rec & Env & Dom & HV & Meta & Acc \\
\midrule
\rowcolor{gray!20}\multicolumn{12}{c}{\textbf{Multilingual}}\\
\midrule
GlobalPhone \cite{schultz02_icslp} & 35 & 170 & 1 & \Xmark & Script & Rec & Ctrl & 1 & \Cmark & \Cmark & \Xmark \\
Common Voice \cite{ardila-etal-2020-common} & 15 & 225 & 1 & \Xmark & Script & Rec & Unctrl & 3 & \Cmark & \Cmark & \Cmark \\
\midrule
\rowcolor{gray!20}\multicolumn{12}{c}{\textbf{Uni-Dialect}}\\
\midrule
FACST \cite{djegdjiga-etal-2018-french} & 7.5 & 20 & 1 & \Xmark & Script & Rec & Ctrl & 1 & \Cmark & \Cmark & \Xmark \\
ArzEn \cite{hamed2021investigationsspeechrecognitionsystems} & 11.4 & 40 & 1 & \Xmark & Conv. & Rec & Ctrl & 2 & \Cmark & \Cmark & \Cmark \\
TUN-SWITCH \cite{abdallah2023leveragingdatacollectionunsupervised} & $\sim$9 & -- & 1 & \Xmark & Broadcast & TV/YT & Mixed & 2 & \Cmark & \Xmark & \Cmark \\
SAAVB \cite{ALGHAMDI200845} & $\sim$96.4 & 1{,}033 & 1 & \Cmark & Script & Tel & Mixed & $\sim$9 & \Cmark & \Cmark & \Cmark \\
ALGASD \cite{DrouaHamdani2010ALGERIANAS} & -- & 300 & 1 & \Xmark & Script & Rec & Ctrl & 1 & \Cmark & \Cmark & \Cmark \\
TARIC \cite{Masmoudi2017AutomaticSR} & $\sim$10 & 108 & 1 & \Xmark & Conv. & Rec & Unctrl & 1 & \Cmark & \Xmark & \Xmark \\
STAC \cite{zribi-etal-2014-conventional} & 5 & $\sim$70 & 1 & \Xmark & Conv. & TV & Ctrl & 5 & \Cmark & \Xmark & \Xmark \\
TunSpeech \cite{Messaoudi2021TunisianDE} & $\sim$10.5 & 10 & 1 & \Xmark & Mixed & YT/TV/Rec & Mixed & 2 & \Cmark & \Xmark & \Xmark \\
\midrule
\rowcolor{gray!20}\multicolumn{12}{c}{\textbf{Multi-Dialect}}\\
\midrule
ESCWA.CS \cite{chowdhury2021modelruleallmultilingual} & 2.8 & -- & 4 & \Xmark & Conv. & Rec & Unctrl & -- & \Cmark & \Xmark & \Xmark \\
OrienTel \cite{siemund2002orientel} & -- & -- & 6 & \Xmark & Conv. & Tel & Mixed & -- & \Xmark & \Xmark & \Cmark \\
Fisher Lev. \cite{Fesharticle} & 45 & -- & 4 & \Xmark & Conv. & Tel & Ctrl & -- & \Xmark & \Xmark & \Xmark \\
QASR \cite{mubarak-etal-2021-qasr} & 2{,}000 & 11{,}092 & 5 & \Xmark & Script & TV & Ctrl & 5 & \Xmark & \Cmark & \Xmark \\
MASC \cite{10022652} & 1{,}000 & -- & 23 & \Xmark & Social & YT & Unctrl & 4 & \Cmark & \Cmark & \Xmark \\
MGB-2 \cite{ali2019mgb2challengearabicmultidialect} & 1{,}200 & -- & 4 & \Xmark & Script & TV & Ctrl & 12 & \Cmark & \Cmark & \Xmark \\
MGB-5 \cite{9003960} & 14 & -- & 17 & \Xmark & Social & YT & Unctrl & 7 & \Cmark & \Cmark & \Xmark \\
\midrule
\textbf{\ourdataset (Ours)} & \textbf{71.59} & \textbf{275} & \textbf{11} & \textbf{\Cmark} & \textbf{Mixed} & \textbf{Rec} & \textbf{Unctrl} & \textbf{11} & \textbf{\Cmark} & \textbf{\Cmark} & \textbf{\Cmark} \\
\bottomrule
\end{tabular*}

\vspace{1mm}
\noindent\textit{Abbreviations:}
Hrs = total hours with transcription; Spk = number of speakers; Dial = number of major dialect groups;
Micro = micro-dialect coverage; Text Src = text prompt source (Script: scripted text; Conv.: conversational speech; Broadcast; Social: social media);
Rec = recording style (Rec: recorded; Tel: telephone conversations; YT: YouTube; TV);
Env = environment (Ctrl: controlled; Unctrl: uncontrolled; Mixed); Dom = number of domains; HV = human verification;
Meta = metadata availability; Acc = accented speech diversity. \Cmark~= yes, \Xmark~= no, -- = not reported.
\end{table*}

Despite progress in Arabic ASR datasets, most of these datasets cover only a few dialects, primarily modern standard Arabic (MSA), Egyptian, broad Levantine, Gulf, and Moroccan \cite{djanibekov2025dialectal}. On the other hand, Algerian, Sudanese, Yemeni, and Tunisian Arabic, as well as parts of Palestinian and Jordanian Arabic, remain comparatively low-resource \cite{sullivan2026arab,alqadasi2025arabic, talafha2024casablanca}. Beyond dialect coverage, existing research has also largely overlooked accented MSA and classical Arabic (CA). MSA is the formal variety used in modern communication, while CA represents the historical form found in religious and classical texts \cite{ryding2005reference,versteegh2014arabic}. To our knowledge, no dedicated speech dataset exists to systematically evaluate accented MSA and CA across diverse dialectal backgrounds in ASR.

To address these limitations, we introduce \ourdataset, a community-driven Arabic speech dataset designed to improve dialectal coverage in ASR, covering~11 countries: \textit{Algeria, Egypt, Jordan, Morocco, Palestine, Saudi Arabia, Sudan, Syria, Tunisia, United Arab Emirates,} and \textit{Yemen}. To capture finer linguistic variation, we further divide the Saudi and Yemeni dialects into fine-grained sub-dialects. The dataset also includes accented MSA and CA, enabling accent-aware evaluation. \ourdataset is constructed from text collected across multiple sources and domains to ensure broad lexical and topical diversity. We employ a two-level human verification process to ensure transcription accuracy, audio quality, and dialect authenticity to improve the dataset's reliability. Moreover, we benchmark a diverse set of existing ASR systems on \ourdataset, establishing comprehensive baselines across dialectal and formal Arabic varieties. We also release demographic metadata, including age and gender, to support demographic analysis and fairness-aware evaluation of ASR models.

\section{Related Work}
\label{sec_related}

\vspace{0mm}
Arabic speech resources span multilingual, uni-dialect, and multi-dialect datasets. Multilingual datasets such as GlobalPhone \cite{schultz02_icslp} and Mozilla Common Voice \cite{ardila-etal-2020-common} include Arabic and support cross-lingual learning, but typically provide limited dialectal diversity. Uni-dialect corpora, including Saudi, Algerian, and Tunisian resources, offer deeper coverage of specific dialects or code-switching scenarios \cite{ALGHAMDI200845,DrouaHamdani2010ALGERIANAS,Masmoudi2017AutomaticSR,Messaoudi2021TunisianDE}, but their narrow geographic focus and controlled recording conditions limit generalizability across the broader Arabic-speaking world.

\vspace{0mm}
Multi-dialect Arabic datasets provide broader linguistic coverage by capturing speech from multiple regions, as seen in OrienTel \cite{siemund2002orientel}, Fisher Levantine \cite{Fesharticle}, QASR \cite{mubarak-etal-2021-qasr}, MGB \cite{ali2019mgb2challengearabicmultidialect}, and MASC \cite{e1qb_jv46_21}. However, many of these datasets rely heavily on broadcast, telephone, or web-sourced recordings, introducing domain bias, inconsistent annotation quality, and limited representation of spontaneous community speech. These limitations highlight the need for more diverse and dialect-rich Arabic speech corpora for robust ASR development. Table~\ref{tab:dataset_comparison_new} compares \ourdataset with existing datasets. Appendix \ref{appendix:related} provides a detailed literature review.

\vspace{0.5mm}
\noindent
\textbf{\ourdataset in Comparison.} Unlike many prior resources that focus on a single dialect, broadcast speech, or controlled scripted recordings, \ourdataset~ captures naturally diverse speech collected from real speakers across multiple dialect families, with explicit sub-dialect annotations for some countries to reflect fine-grained regional variation.
Furthermore, \ourdataset~ distinguishes itself by including accented MSA and CA recordings from 11 dialect speakers, enabling accent-aware ASR analysis under standardized linguistic content. Combined with human-verified transcriptions, rich demographic metadata, and balanced domain diversity, this makes \ourdataset~a more realistic and comprehensive benchmark for developing robust Arabic ASR systems.

\section{\ourdataset~Dataset}
\label{sec_dataset}
\begin{figure*}[t]
    \centering
    \includegraphics[width=\textwidth]{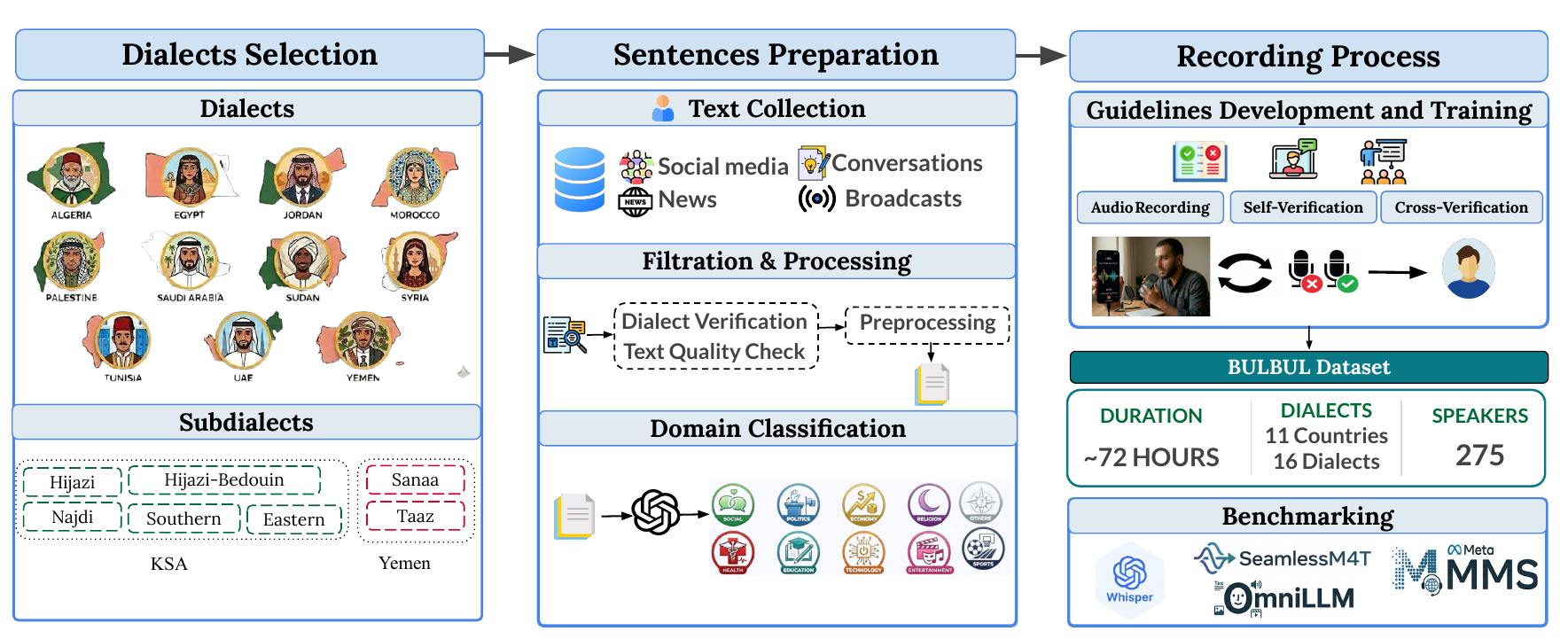}
    \caption{\ourdataset~dataset construction pipeline. }
    \label{fig:pipline}
\end{figure*}
The \ourdataset~dataset is a community-driven project conducted from January 2025 to April 2026. The participants comprise 275~members from~11 Arab countries, of whom~160 were male and~115 were female.  We organized the participants into country-specific teams, each headed by a leader. The leaders were responsible for collecting and verifying the raw text, providing guidance to participants during recording, and verifying the recorded audio. To ensure steady progress, we maintained continuous communication through a dedicated Slack group for discussions and held weekly meetings to review progress, present statistics, and address any challenges. Figure~\ref{fig:pipline} illustrates the pipeline for curating the \ourdataset~dataset.

\vspace{-1mm}
\subsection{Data Collection}



The dataset collection process was conducted incrementally. It initially began with two countries, Saudi Arabia and Egypt. After establishing the collection pipeline and validation procedures, additional countries were gradually incorporated, bringing the total to~11: \textit{Algeria (AG), Egypt (EG), Jordan (JO), Morocco (MA), Palestine (PS), Saudi Arabia (KSA), Sudan (SD), Syria (SY), Tunisia (TN), United Arab Emirates (UAE)}, and \textit{Yemen (YE)}. The resulting dataset includes dialectal Arabic recordings and formal Arabic speech, comprising CA and MSA spoken in participants’ native dialectal accents. Therefore, \ourdataset enables the study of dialectal and accent-aware speech recognition.

We used the \textit{white dialect} {\footnotesize \setcode{utf8}\< اللهجة البيضاء >} for seven countries: \textit{Algeria, Egypt, Jordan, Morocco, Palestine, Sudan,} and \textit{Tunisia}. The \textit{white dialect} is a neutralized variety that blends regional dialectal features with widely understood MSA while avoiding highly localized vocabulary \cite{alkhamees2023white}. It is commonly used in media, cross-regional communication, and formal social contexts within each country.
For Syria and the United Arab Emirates, we focused on the \textit{Levantine} {\footnotesize \setcode{utf8}\< الشامية >} and \textit{Abu Dhabi} {\footnotesize \setcode{utf8}\< الظبيانية >} dialects, respectively.

For Saudi and Yemeni dialects, we collected data at the sub-dialect\footnote{Sub-dialect: a regional variety within a country.} level to better capture intra-country variation. These dialects were selected based on the availability of sub-dialectal data and speaker contributions from these regions. 
The Saudi dialect is divided into six sub-dialects: \textit{Hijazi} (HJ) {\footnotesize \setcode{utf8}\< حجازية >}, \textit{Qatifi} (QT) {\footnotesize \setcode{utf8}\< قطيفية >},
\textit{Southern} (ST) {\footnotesize \setcode{utf8}\< جنوبية >},
\textit{Najdi} (NJ) {\footnotesize \setcode{utf8}\< نجدية >}, \textit{Hijazi-Badwi} (HJ-BD) {\footnotesize \setcode{utf8}\< حجازية بدوية >}, and \textit{Hassawi} (HS) {\footnotesize \setcode{utf8}\< حساوية >}. We combined QT and HS to represent the \textit{Eastern} (ES) dialect. For Yemen, we collected data from two sub-dialects: \textit{Sana'ani} (SN) {\footnotesize \setcode{utf8}\< صنعاني >} and \textit{Ta’izzi} (TZ) {\footnotesize \setcode{utf8}\< تعزي >}. 

\vspace{0.3mm}\noindent
\textbf{Text Collection.} 
The text used in our dataset was collected separately for each Arabic dialect to ensure adequate representation of linguistic diversity across regions. For each dialect, texts were collected from multiple sources, including publicly available datasets (e.g., Casablanca \cite{talafha2024casablanca}, SADA \cite{alharbi2024sada}, and MADAR \cite{bouamor2018madar}), social media platforms (e.g., Facebook, X, and YouTube comments), and private communication via platforms, such as WhatsApp and Telegram. This multi-source collection strategy was designed to capture naturalistic, regionally grounded linguistic variation. For accented MSA and CA, we used texts from the Targamat dataset~\cite{kadaoui2023tarjamat}, consisting of 200 CA samples and 200 MSA samples. In Appendix \ref{semantic}, we provide a detailed lexical and semantic analysis of the collected textual data.

\vspace{0.3mm}\noindent
\textbf{Text Revision.} After the initial text collection stage, we manually reviewed the text to remove offensive and slang content, politically sensitive material, and texts that are not aligned with the targeted dialects. For the tarjamat dataset, the filtration resulted in 129 MSA and 200 CA sentences. The reviewed texts were then passed to a pre-processing pipeline before being uploaded to the recording tool, as described in Appendix \ref{process}.

\vspace{1mm}\noindent
\textbf{Domain Classification.}
We employed GPT-5-mini to categorize the collected texts into thematic domains. The model was prompted with a structured domain taxonomy, consisting of~11 predefined categories (Social, Politics, Economy, Religion, Health, Education, Technology, Entertainment, Sports, Crime, and Other), along with concise definitions for each domain. Appendix \ref{domains} provides details about the domain classification process and human review. 
















\vspace{-0.5mm}
\subsection{Recording Process}
The collected texts from all dialects were uploaded to a recording tool developed using Gradio\footnote{https://www.gradio.app/}. We organized the text by country and, if applicable, sub-dialect. We provided participants with written guidelines (Appendix \ref{recording}) to guide them through the recording stage.
We also recorded an instructional video to help participants become familiar with the recording tool and to minimize the risk of errors. 
Participants were recruited using different strategies depending on the country. In Saudi Arabia, volunteers were recruited through an online volunteering opportunity published on the Saudi National Volunteer Platform\footnote{Saudi NVG: 
\url{https://nvg.gov.sa/home}}. For every 10 minutes of recorded audio, participants were credited with one hour of voluntary work. For other contributing countries, the participants were selected by country leaders.

\vspace{-1mm}
\subsection{Human Sanity Check}
We used two levels of revision. The first level is self-verification, in which the speaker optionally listens to their own recording before submission to ensure clarity and alignment with the text. The second level is external verification, in which a native member of the corresponding country reviews the recordings. We provided the reviewers with written guidelines and a video recording explaining the verification strategy. The detailed verification rules are outlined in Appendix \ref{recording}.

\section{\ourdataset~Analysis}
\label{sec:analysis}

\textbf{Overall Dataset Distribution.} 
Tables~\ref{tab:dataset_stats} and~\ref{tab:msa_ca_stats} show the statistics of the \ourdataset~dataset.
\ourdataset consists of 36,949 dialectal utterances recorded over 61.46 hours alongside 2,325 accented MSA/CA recordings totaling 10.42 hours. 
The longest recordings in the dialectal subset are for the Yemeni (18.97 hours), Saudi (11.89 hours), and Syrian (8.21 hours) dialects. In contrast, the Sudanese dialect remains comparatively underrepresented, with less than one hour of recorded speech. This lower representation is mainly due to the limited availability of native Sudanese participants during the data collection period.

In the accented MSA/CA subset, the highest durations are observed for Yemeni (2.82 hours) and Egyptian (2.02 hours) dialects, while the lowest are observed for UAE (0.02 hours) and Tunisian (0.11 hours) dialects. This imbalance highlights both the practical challenges of large-scale dialectal data collection and the need for continued efforts to expand coverage of lower-resourced varieties.
\\
\noindent
\textbf{Sub-dialectal Distribution.} As shown in Table~\ref{tab:dataset_stats}, the two countries, further divided into sub-dialects, are Saudi Arabia and Yemen. In Saudi Arabia, the largest number of recordings is for the Hijazi dialect, with 7,120 utterances totaling 10,73 hours. In Yemen, the Ta'izzi dialect contains nearly twice as many utterances and approximately 1.6 times as many speech hours as the Sana'ani dialect.

\begin{table}[t]
\centering
\footnotesize
\setlength{\tabcolsep}{4pt}
\caption{Descriptive statistics \ourdataset across countries. We report the total number of utterances (Utts), total hours, minimum duration (Min) in seconds, average duration (AvgDur) in seconds, maximum duration (Max) in seconds, average words per utterance (WPU), and average words per second (WPS). Saudi Arabia and Yemen are further divided into subdialects.}
\label{tab:dataset_stats}

\sisetup{
  round-mode=places,
  round-precision=2,
  table-number-alignment=center
}

\begin{adjustbox}{max width=\linewidth}
\begin{tabular}{
l
r
S[table-format=2.2]
S[table-format=1.3]
S[table-format=1.3]
S[table-format=3.3]
S[table-format=2.2]
S[table-format=1.2]
}
\toprule
\textbf{Dialect} & \textbf{Utts} & {\textbf{Hrs}} & {\textbf{Min}} & {\textbf{AvgDur}} & {\textbf{Max}} & {\textbf{WPU}} & {\textbf{WPS}} \\
\midrule

AG   & 1,172 & 2.21  & 2.100 & 6.781 & 120.420 & 10.06 & 1.48 \\
EG     & 2,034 & 3.09  & 1.74 & 5.462 & 34.38  & 10.58 & 1.94 \\
JO    & 3,284 & 4.92  & 1.721 & 5.392 & 23.460  & 9.86  & 1.83 \\
\midrule
KSA & 7,872 & 11.89 & 0.834 & 5.440 & 152.340 & 9.54  & 1.75 \\
\quad HJ         & 7,120 & 10.73 & 1.023 & 5.426 & 152.340 & 9.51  & 1.75 \\
\quad HJ-BD & 255   & 0.49  & 2.908 & 6.911 & 30.398  & 11.35 & 1.64 \\
\quad ST       & 307   & 0.56  & 2.353 & 6.560 & 29.548  & 9.66  & 1.47 \\
\quad NJ          & 761   & 0.85  & 1.820 & 4.024 & 10.260  & 8.72  & 2.17 \\
\quad ES        & 449   & 1.01  & 0.834 & 8.127 & 25.260  & 14.31 & 1.76 \\
\midrule
MA   & 1,169 & 2.30  & 2.520 & 7.096 & 38.804  & 11.22 & 1.58 \\
PS & 2,444 & 3.63  & 1.140 & 5.347 & 47.760  & 8.27  & 1.55 \\

SD   & 577   & 0.80 & 1.920 & 4.983 & 16.740 & 6.68 & 1.34 \\
SY   & 5,449 & 8.21 & 1.500 & 5.427 & 50.820 & 7.28 & 1.34 \\
TN & 1,959 & 3.34 & 1.181 & 6.138 & 22.980 & 8.32 & 1.36 \\
UAE     & 1,518 & 2.10 & 1.721 & 4.969 & 39.807 & 8.02 & 1.61 \\
\midrule
YE       & 9,470 & 18.97 & 1.014 & 7.211 & 348.960 & 7.85 & 1.09 \\
\quad SN   & 3,145 & 7.38  & 1.380 & 8.450 & 348.960 & 9.97 & 1.18 \\
\quad TZ    & 6,322 & 11.58 & 1.014 & 6.596 & 300.600 & 6.79 & 1.03 \\

\midrule
\textbf{Summary} & \textbf{36,949} & \textbf{61.46} & \textbf{0.834} & \textbf{6.00} & \textbf{348.960} & \textbf{8.80} & \textbf{1.46} \\
\bottomrule
\end{tabular}
\end{adjustbox}
\end{table}





\begin{table}[t!]
\centering
\scriptsize
\setlength{\tabcolsep}{2pt}
\caption{Descriptive statistics of the accented MSA and CA subset across dialect groups. Dial denotes dialects, and Utts denotes the number of utterances.}
\label{tab:msa_ca_stats}
\sisetup{
  round-mode=places,
  round-precision=2,
  table-number-alignment=center
}

\begin{adjustbox}{max width=\linewidth}
\begin{tabular}{
l
r
S[table-format=2.2]
S[table-format=2.3]
S[table-format=2.3]
S[table-format=2.3]
S[table-format=2.2]
S[table-format=1.2]
}
\toprule
\textbf{Dial} & \textbf{Utts} & {\textbf{Hrs}} & {\textbf{Min}} & {\textbf{Avg}} & {\textbf{Max}} & {\textbf{WPU}} & {\textbf{WPS}} \\
\midrule

AG  & 68  & 0.35 & 8.280  & 18.722 & 40.320 & 32.68 & 1.75 \\
EG  & 447 & 2.02 & 4.800  & 16.273 & 61.320 & 26.52 & 1.63 \\
JO  & 67  & 0.28 & 1.920  & 15.017 & 40.430 & 28.96 & 1.93 \\
KSA & 102 & 0.40 & 1.260  & 14.132 & 37.488 & 24.47 & 1.73 \\
MA  & 55 & 0.29   & 17.58    & 19.18    & 48.00    & 27.82   & 1.45  \\
PS  & 418 & 1.78 & 4.500  & 15.357 & 47.390 & 26.06 & 1.70 \\
SD  & 80  & 0.41 & 7.195  & 18.659 & 41.520 & 28.38 & 1.52 \\
SY  & 389 & 1.94 & 6.000  & 17.908 & 46.620 & 26.44 & 1.48 \\
TN  & 22  & 0.11 & 11.700 & 17.752 & 26.700 & 29.18 & 1.64 \\
UAE & 6   & 0.02 & 6.240  & 10.420 & 16.140 & 16.50 & 1.58 \\
YE  & 671 & 2.82 & 3.780  & 15.153 & 48.780 & 26.47 & 1.75 \\

\midrule
\textbf{Summary} & \textbf{2,325} & \textbf{10.42} & \textbf{1.260} & \textbf{16.234} & \textbf{61.320} & \textbf{26.68} & \textbf{1.65} \\
\bottomrule
\end{tabular}
\end{adjustbox}
\end{table}

\vspace{0mm}
\vspace{-0.5mm}
\noindent
\textbf{Cross-dialectal Speech Tempo.}
To directly assess speaking tempo, we analyze the WPS, which normalizes the utterance length by duration. Clear cross-dialectal variation is observed in the dialectal subset, with Egyptian dialect exhibiting the fastest speech tempo (1.94 WPS), followed by Jordanian (1.83 WPS), while Yemeni shows the slowest tempo (1.09 WPS). Interestingly, accented MSA/CA speech exhibits a higher average tempo (1.65 WPS) than dialectal speech (1.46 WPS), suggesting faster, more fluent articulation in the scripted formal setting. This contrast is particularly pronounced for Yemeni speakers, whose dialectal speech has the slowest tempo yet becomes among the fastest in the accented subset (1.75 WPS). These findings suggest that speaking style, recording conditions, and linguistic structure all influence temporal speech characteristics. 
The observed differences between dialectal speech and MSA/CA should be interpreted with caution, as the MSA/CA subset was recorded under different conditions, including read speech, which may influence speaking rate.
From an ASR perspective, such tempo variation is important, as faster speech may challenge temporal alignment and acoustic modeling, whereas slower speech may introduce longer temporal dependencies \cite{talafha2024casablanca}.

\vspace{0.5mm}
\noindent
\textbf{Outlier Analysis.}
As shown in Table \ref{tab:dataset_stats}, the dataset exhibits noticeable variability in maximum utterance duration across dialects, with some recordings substantially exceeding the typical range. For example, Yemeni Arabic has a maximum recording duration of 348,96 seconds, while the Saudi dialect reaches 152.340 seconds, both significantly higher than the average segment duration across dialects. Such long recordings likely represent long scripts.
\begin{figure}[t!]
    \centering
    \includegraphics[width=\linewidth]{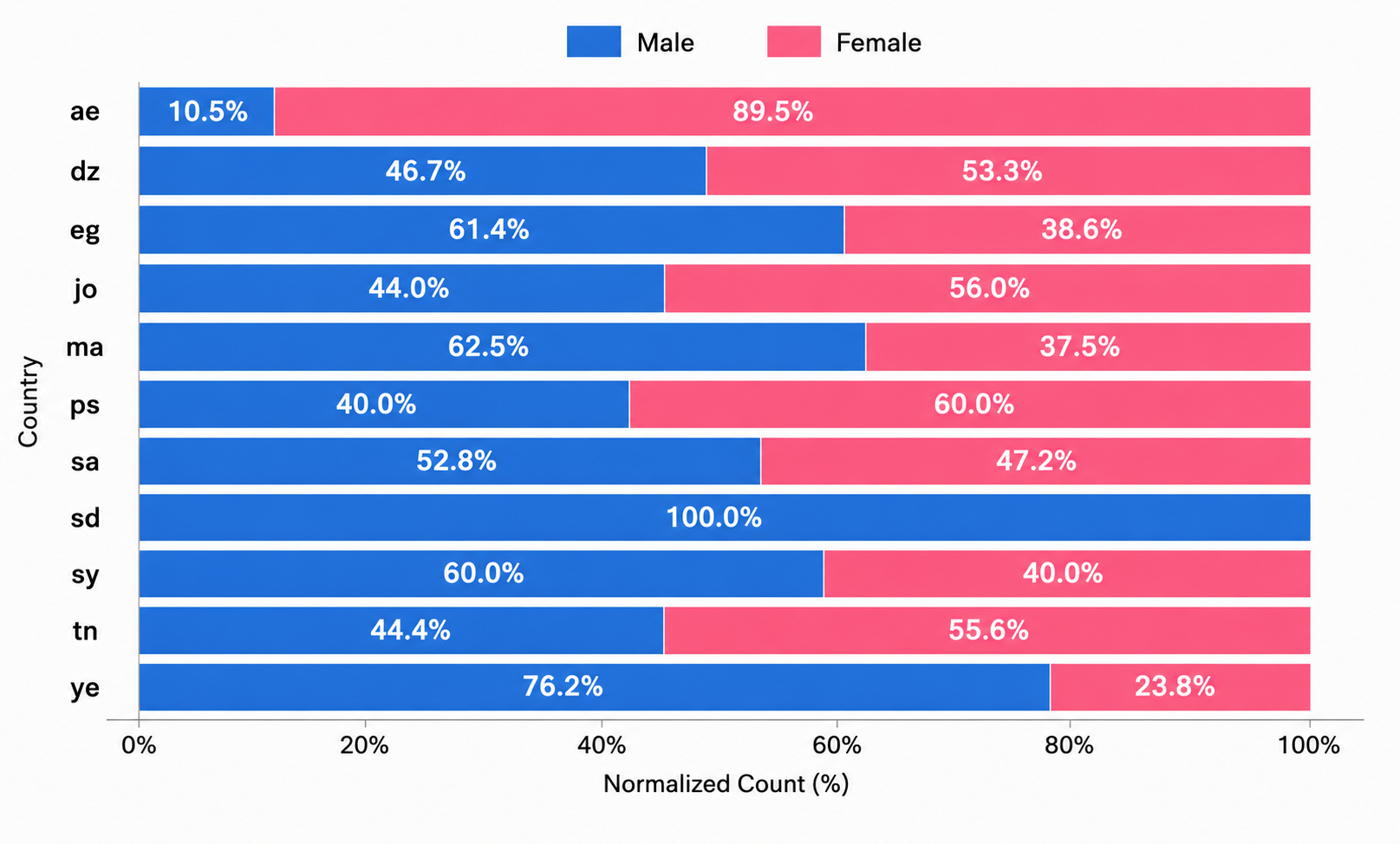}
    \caption{Normalized distribution of male and female speakers across the represented countries.}
    \label{fig:gender}
\end{figure}

\vspace{0.5mm}
\noindent
\textbf{Gender Distribution.}
\ourdataset~exhibits variability in gender representation across countries, as shown in Figure \ref{fig:gender}. Approximately balanced gender distributions are observed in Algeria ($53.3\%$ female), Jordan ($56.0\%$ female), Saudi Arabia ($52.8\%$ male), and Tunisia ($55.6\%$ female). Other countries, such as Sudan, Yemen, Algeria, and the UAE, show skewed distributions. Sudan includes only male participants, while the UAE shows a strong female majority ($\sim$89.5\%). Although the recruitment strategy aimed to encourage diversity, these were recruitment objectives rather than enforced demographic quotas. As \ourdataset is a community-driven dataset relying on voluntary participation and participant availability, achieving an equal gender distribution was not always possible across all countries. 

\begin{figure}[t!]
    \centering
    \includegraphics[width=\linewidth]{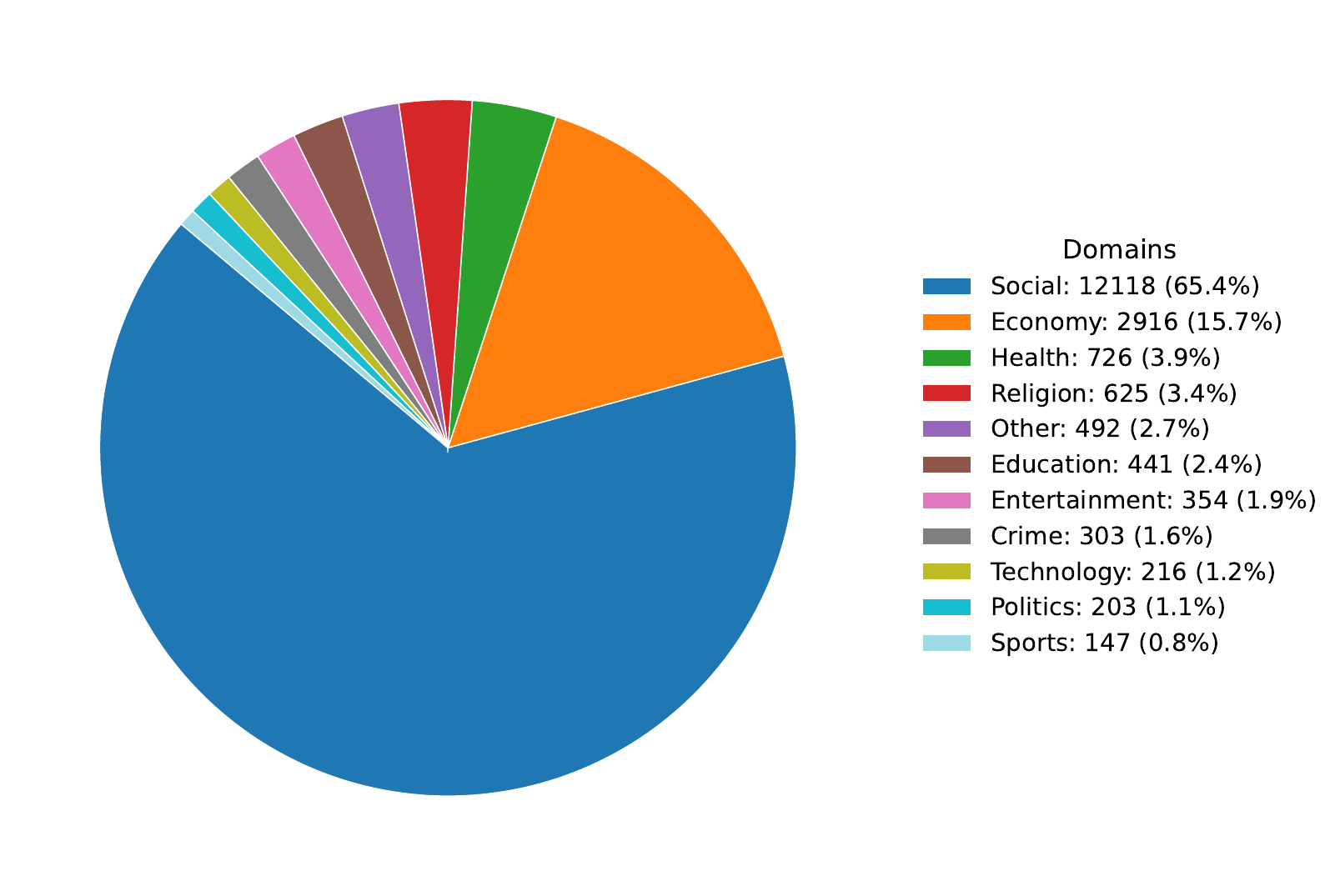}
    \caption{Overall domain distribution across the 11 participating countries.}
    
    \label{fig:domains}
\end{figure}

\vspace{0.5mm}
\noindent
\textbf{Domain Distribution.}
To ensure topical diversity, we analyzed the domain composition of the collected corpus across the 11 participating countries. Figure \ref{fig:domains} shows the overall domain distribution, revealing a strong concentration in the \textit{Social} domain, which constitutes the largest portion of the dataset (65.4\%), followed by \textit{Economy} (15.7\%). The remaining domains are more sparsely represented, each contributing a smaller proportion. This imbalance reflects the natural prevalence of conversational and socially oriented content in community-contributed speech data, while still maintaining broad coverage across topics.




\vspace{0.5mm}
\noindent
\textbf{Data Splits.}
To ensure a robust and fair evaluation of ASR models' performance on the \ourdataset~dataset, we rely on the human-verified subset for development and testing. The total duration of verified data varies across countries. Thus, instead of enforcing a fixed absolute duration per split, we divide the verified data within each country into 50\% development and 50\% test splits that are speaker-disjoint.  Table~\ref{tab:split_stats} reports the statistics of the development and test splits for each country, including the number of utterances, total duration, and number of speakers. 

\section{Evaluation}
\label{sec_evaluation}

\textbf{Evaluated Models}
To benchmark \ourdataset, we evaluate a diverse set of state-of-the-art multilingual speech recognition models spanning a wide range of model scales, including \textit{Whisper Large-V3} \cite{radford2023robust}, \textit{Whisper Large-V3-Turbo} \cite{radford2023robust}, \textit{SeamlessM4T} \cite{barrault2023seamlessm4t}, \textit{massively multilingual speech} \textit{(MMS)} \cite{pratap2024scaling}, \textit{OmniLLM} \cite{omnilingual2025omnilingual} (300M, 1B, 3B, and 7B), and \textit{OmniCTC} \cite{omnilingual2025omnilingual} (300M, 1B, 3B, and 7B).

\vspace{0.5mm}
\noindent
\textbf{Evaluation Setup}
To evaluate the models on \ourdataset, we used a zero-shot evaluation setup using the test set without fine-tuning on the \ourdataset~dataset or any country-specific subset. The evaluation is conducted using each model’s default decoding configuration. These settings allow us to measure out-of-the-box generalization performance and assess both cross-lingual and cross-accent robustness.
More details are present in Appendix \ref{setup}.

\noindent
\textbf{Evaluation Protocol} To ensure that the reported character error rate (CER) and word error rate (WER) accurately reflect transcription quality, the same text normalization pipeline was applied to both the reference and predicted transcripts. The details are described in Section~\ref{text_norm}.

\section{Results and Discussion}
\label{sec:results}
We evaluate model performance on the \ourdataset test set using WER and CER. Tables~\ref{tab:wer_cer} and~\ref{tab:msa_wer_cer} report the results across all evaluated systems tested on dialectal subsets and accented MSA/CA, respectively. For analysis, we group countries into broader regional dialect categories: Northwest African, Arabian Peninsula, Levantine, and Nile Valley, based on geographic proximity and established Arabic dialect classifications.

\definecolor{rankone}{RGB}{180,230,180}
\definecolor{ranktwo}{RGB}{255,255,255}
\definecolor{worst}{RGB}{255, 228, 225}

\begin{table*}[t!]
\centering
\caption{Performance of ASR models on dialectal Arabic speech of \ourdataset~dataset grouped by regional dialect clusters. Each entry is reported as WER/CER (\%).
\colorbox{rankone}{\phantom{X}} is the lowest and 
\colorbox{worst}{\phantom{X}} is the highest WER/CER per dialect.
}

\renewcommand{\arraystretch}{1.15}
\resizebox{\linewidth}{!}{%
\begin{tabular}{l|ccc|ccc|ccc|cc|c}
\toprule
\textbf{Model} 
& \multicolumn{3}{c|}{\textbf{Northwest African }} 
& \multicolumn{3}{c|}{\textbf{Arabian Peninsula}} 
& \multicolumn{3}{c|}{\textbf{Levantine}} 
& \multicolumn{2}{c|}{\textbf{Nile Valley}}
& \textbf{Mean} \\
\cmidrule(lr){2-4} \cmidrule(lr){5-7} \cmidrule(lr){8-10} \cmidrule(lr){11-12}

& Algeria & Tunisia & Maghrib
& UAE & KSA & Yemen
& Jordan & Syria & Palestine
& Egypt & Sudan \\
\midrule

WhisperV3
& 80.5/42.1 & 48.4/11.7 & 71.7/33.1
& 46.8/13.9 & 26.0/9.7 & 67.9/31.7
& 34.9/8.6 & 58.9/17.9 & 29.6/9.3
& 61.6/29.8 & 65.6/38.5
& 53.8/22.9 \\

WhisperV3T
& 88.2/\colorbox{worst}{46.0} & 51.2/12.7 & 67.5/25.2
& 60.2/19.9 & 33.3/12.0 & 79.8/35.3
& 39.6/10.7 & 75.2/28.4 & 34.7/13.0
& 81.6/\colorbox{worst}{44.3} & 83.8/41.7
& 63.2/26.8 \\

Seamless
& 69.5/28.0 & 45.2/10.6 & 56.1/21.6
& 45.9/12.7 & 27.4/10.3 & 62.9/22.5
& 34.4/8.1 & 60.1/17.9 & 26.1/7.2
& 43.3/\colorbox{ranktwo}{16.3} & 70.9/30.1
& 49.8/16.6 \\

MMS1B
& \colorbox{worst}{88.5}/31.7 & \colorbox{worst}{70.7}/18.8 & \colorbox{worst}{83.8}/29.5
& \colorbox{worst}{72.9}/21.0 & \colorbox{worst}{66.8/22.3} & \colorbox{worst}{82.3}/30.2
& \colorbox{worst}{71.9/20.8} & \colorbox{worst}{84.4}/28.1 & \colorbox{worst}{62.4/18.8}
& \colorbox{worst}{84.9}/34.0 & 79.1/27.5
& \colorbox{worst}{77.1}/25.1 \\

OmniLLM300M
& 66.4/23.8 & 46.7/11.4 & 54.0/16.3
& 53.3/16.7 & 27.0/8.7 & 62.5/22.0
& 35.5/8.6 & 63.2/20.9 & 24.8/7.4
& 51.2/20.1 & 68.5/27.0
& 50.8/17.0 \\

OmniLLM1B
& 60.0/19.2 & \colorbox{ranktwo}{43.8}/10.5 & 49.1/14.5
& 46.5/13.3 & 22.3/\colorbox{ranktwo}{7.0} & 57.1/17.7
& 33.3/\colorbox{ranktwo}{7.7} & 58.4/18.3 & \colorbox{ranktwo}{19.8/5.4}
& 45.1/17.0 & 61.0/22.5
& 45.1/13.9 \\

OmniLLM3B
& \colorbox{ranktwo}{58.1}/19.1 & \colorbox{rankone}{42.7/8.9} & \colorbox{ranktwo}{48.0/13.9}
& \colorbox{ranktwo}{43.6/11.8} & \colorbox{ranktwo}{19.7}/\colorbox{rankone}{6.5} & \colorbox{ranktwo}{56.9}/\colorbox{rankone}{16.8}
& \colorbox{ranktwo}{31.6}/\colorbox{rankone}{6.8} & \colorbox{ranktwo}{56.2/16.8} & 20.0/5.5
& \colorbox{ranktwo}{43.1}/16.8 & \colorbox{ranktwo}{59.3/20.5}
& \colorbox{ranktwo}{43.6/13.0} \\

OmniLLM7B
& \colorbox{rankone}{56.1}/\colorbox{ranktwo}{18.4}
& \colorbox{rankone}{42.7}/\colorbox{ranktwo}{10.0}
& \colorbox{rankone}{45.3/12.2}
& \colorbox{rankone}{43.0/10.9}
& \colorbox{rankone}{19.6/6.5}
& \colorbox{rankone}{54.5}/\colorbox{ranktwo}{17.8}
& \colorbox{rankone}{30.8/6.8}
& \colorbox{rankone}{52.1/14.9}
& \colorbox{rankone}{17.8/4.8}
& \colorbox{rankone}{40.2/14.3}
& \colorbox{rankone}{56.5/19.1}
& \colorbox{rankone}{41.7/12.7} \\

OmniCTC300M
& 76.0/28.6 & 57.7/18.6 & 81.5/\colorbox{worst}{53.4}
& 66.0/\colorbox{worst}{25.9} & 49.3/21.6 & 75.9/\colorbox{worst}{36.3}
& 51.5/15.0 & 76.2/\colorbox{worst}{30.1} & 39.0/12.0
& 69.2/30.4 & \colorbox{worst}{84.0}/\colorbox{worst}{44.2}
& 65.7/\colorbox{worst}{28.8} \\

OmniCTC1B
& 58.5/\colorbox{rankone}{15.9}& 46.2/12.2 & 60.0/20.6
& 47.2/13.7 & 26.9/7.9 & 62.6/23.8
& 39.0/9.7 & 60.2/20.2 & 26.0/7.6
& 52.6/20.3 & 65.8/28.7
& 49.5/16.6 \\

OmniCTC3B
& 71.1/42.1 & 52.4/19.5 & 68.5/39.8
& 51.5/19.6 & 30.0/10.3 & 68.4/32.9
& 38.8/10.9 & 65.2/27.5 & 24.3/6.5
& 51.6/20.1 & 70.3/32.2
& 53.3/23.8 \\

OmniCTC7B
& 66.3/30.4 & 51.1/\colorbox{worst}{19.8} & 65.9/38.3
& 49.4/17.4 & 28.4/10.1 & 67.3/30.1
& 36.3/10.2 & 61.8/24.4 & 23.9/6.6
& 50.1/20.5 & 71.4/37.1
& 51.8/22.1 \\

\bottomrule
\end{tabular}
}

\label{tab:wer_cer}
\end{table*}

\begin{table*}[t!]
\centering
\caption{Performance of ASR models on accented MSA/CA speech of \ourdataset~dataset grouped by regional accent clusters. Each entry is reported as WER/CER (\%). 
\colorbox{rankone}{\phantom{X}} is the lowest and 
\colorbox{worst}{\phantom{X}} is the highest WER/CER per dialect.}
\resizebox{\linewidth}{!}{%
\begin{tabular}{l|ccc|ccc|ccc|cc|c}
\toprule
\textbf{Model}
& \multicolumn{3}{c|}{\textbf{Northwest African}}
& \multicolumn{3}{c|}{\textbf{Arabian Peninsula}}
& \multicolumn{3}{c|}{\textbf{Levantine}}
& \multicolumn{2}{c|}{\textbf{Nile Valley}}
& \textbf{Mean} \\
\cmidrule(lr){2-4} \cmidrule(lr){5-7} \cmidrule(lr){8-10} \cmidrule(lr){11-12}
& Algeria & Tunisia & Maghrib
& UAE & KSA & Yemen
& Jordan & Syria & Palestine
& Egypt & Sudan & \\
\midrule

WhisperV3
& 15.4/6.0 & 24.1/6.7 & 18.4/7.1
& 13.3/3.0 & 15.2/4.4 & 14.3/5.5
& 17.0/4.5 & 12.4/5.8 & 15.3/6.4
& 19.0/6.9 & 34.2/\colorbox{worst}{15.3}
& 19.2/8.0 \\

WhisperV3T
& 16.4/6.4 & 25.9/7.2 & 19.7/7.3
& 11.2/2.6 & 16.1/4.7 & 15.8/5.8
& 16.1/3.9 & 11.8/\colorbox{worst}{5.1} & 14.9/6.2
& 19.4/6.8 & 36.2/15.2
& 20.0/7.5 \\

MMS1B
& \colorbox{worst}{30.6/7.5} & \colorbox{worst}{45.1/12.3} & \colorbox{worst}{36.6/9.6}
& \colorbox{worst}{21.4/5.8} & \colorbox{worst}{27.1/7.3} & \colorbox{worst}{25.6/6.3}
& \colorbox{worst}{36.2/8.9} & \colorbox{worst}{23.6/5.1} & \colorbox{worst}{30.7/7.3}
& \colorbox{worst}{34.4/8.8} & \colorbox{worst}{44.8}/13.7
& \colorbox{worst}{33.8/9.7} \\

Seamless
& 11.2/2.9 & \colorbox{rankone}{17.1}/5.1 & \colorbox{rankone}{12.1}/3.7
& \colorbox{rankone}{08.2/1.9} & 16.9/5.1 & 10.7/3.0
& 16.5/4.1 & 9.2/2.3 & 10.9/3.2
& 13.1/3.6 & 23.2/7.7
& 14.4/5.4 \\

OmniLLM300M
& 12.8/3.2 & 23.3/7.1 & 14.5/4.0
& 12.2/3.4 & 14.5/3.9 & 13.0/3.8
& 18.3/4.7 & 9.1/2.2 & 13.4/3.8
& 17.7/5.0 & 27.6/9.0
& 17.0/6.1 \\

OmniLLM1B
& 10.1/2.4 & 19.9/5.7 & 12.8/3.2
& 10.2/2.4 & 13.5/4.0 & 11.9/3.4
& 13.5/3.1 & \colorbox{rankone}{7.8}/1.8 & 12.3/3.4
& 14.4/3.9 & 23.4/7.8
& 14.2/5.3 \\

OmniLLM3B
& 10.2/2.7 & 18.0/\colorbox{rankone}{5.0} & 13.2/\colorbox{rankone}{3.0}
& 11.2/2.6 & 13.5/4.1 & 10.2/2.9
& 13.1/3.1 & 7.9/2.0 & 10.6/3.1
& 14.4/3.9 & 22.1/7.1
& 13.7/5.1 \\

OmniLLM7B
& \colorbox{rankone}{08.9/2.1} & 18.5/5.2 & 12.9/\colorbox{rankone}{3.0}
& 12.2/2.8 & \colorbox{rankone}{13.0/3.7} & \colorbox{rankone}{9.4/2.5}
& 12.3/\colorbox{rankone}{2.7} & 8.1/\colorbox{rankone}{1.7} & \colorbox{rankone}{10.2}/ 2.5
& \colorbox{rankone}{12.7/3.5} & \colorbox{rankone}{21.6/6.3}
& \colorbox{rankone}{13.3/4.8} \\

OmniCTC300M
& 21.3/5.0 & 37.6/9.9 & 27.9/7.5
& 17.4/3.9 & 20.9/5.3 & 19.3/4.7
& 26.4/6.3 & 15.5/3.2 & 22.5/5.3
& 29.9/7.9 & 39.1/12.5
& 26.9/7.8 \\

OmniCTC1B
& 13.4/3.2 & 26.1/7.9 & 18.7/4.5
& 14.3/3.6 & 14.3/3.9 & 13.1/3.0
& 16.7/3.8 & 10.3/2.3 & 14.1/3.9
& 18.4/4.8 & 28.5/8.7
& 18.3/6.0 \\

OmniCTC3B
& 11.8/2.7 & 22.4/6.0 & 14.9/3.7
& 11.2/2.8 & 14.2/3.9 & 11.6/2.8
& 14.6/3.2 & 9.1/2.0 & 11.7/2.8
& 15.9/4.0 & 24.5/7.2
& 15.3/5.2 \\

OmniCTC7B
& 10.6/2.5 & 21.7/5.3 & 14.8/3.5
& 14.3/3.4 & 14.3/\colorbox{rankone}{3.7} & 11.0/\colorbox{rankone}{2.5}
& \colorbox{rankone}{12.0/2.7} & 9.0/2.0 & 10.9/\colorbox{rankone}{2.4}
& 14.3/3.9 & 24.3/6.8
& 15.4/5.1 \\

\bottomrule
\end{tabular}
}
\label{tab:msa_wer_cer}
\end{table*}

\vspace{1mm}
\noindent
\textbf{Overall Results.}
Overall, Omni-based models consistently outperform the baseline multilingual ASR models across regional dialect clusters, with \textit{OmniLLM-7B} achieving the best overall performance in dialectal speech, with a mean WER of 41.7 and a CER of 12.7. This is because OmniLLM-7B processes acoustic features through an LLM decoder, enabling stronger contextual understanding during transcription. This architecture likely improves the model's ability to select contextually appropriate words and generalize more effectively to previously unseen regional Arabic dialects in zero-shot settings.

In the MSA/CA data, performance improves substantially across all models, and \textit{OmniLLM-7B} again achieves the best results. This suggests that dialectal variation poses a greater challenge to multilingual ASR systems than standardized speech. 

\vspace{1mm}
\noindent
\textbf{Per-dialect Results.}
The performance varies substantially across dialects. The Palestinian and Saudi dialects consistently achieve the lowest error rates across most models, indicating comparatively stronger recognition performance. In particular, the Palestinian subset yields the lowest WER for 9 out of the 12 evaluated systems. This is likely driven by the limited domain diversity of the Palestinian subset, which reduces lexical and semantic variability, resulting in samples that is easier for ASR models to transcribe. 

In contrast, Algeria and Sudan exhibit the highest error rates across most models, reflecting greater recognition difficulty in these dialects. For Sudanese Arabic, this may be partly attributed to the limited availability of public speech datasets. On the other hand, Algerian Arabic likely poses greater recognition difficulties due to its substantial lexical and phonological divergence from MSA, highly non-standard orthographic conventions, and the spontaneous conversational nature of the speech, which leads to a stronger mismatch with multilingual ASR training distributions. Moreover, the relatively long average utterance duration in the Algerian subset may further contribute to increased transcription difficulty, as longer sequences are more susceptible to cumulative decoding errors and context drift.

\begin{figure}[t!]
    \centering
    \includegraphics[width=\linewidth]{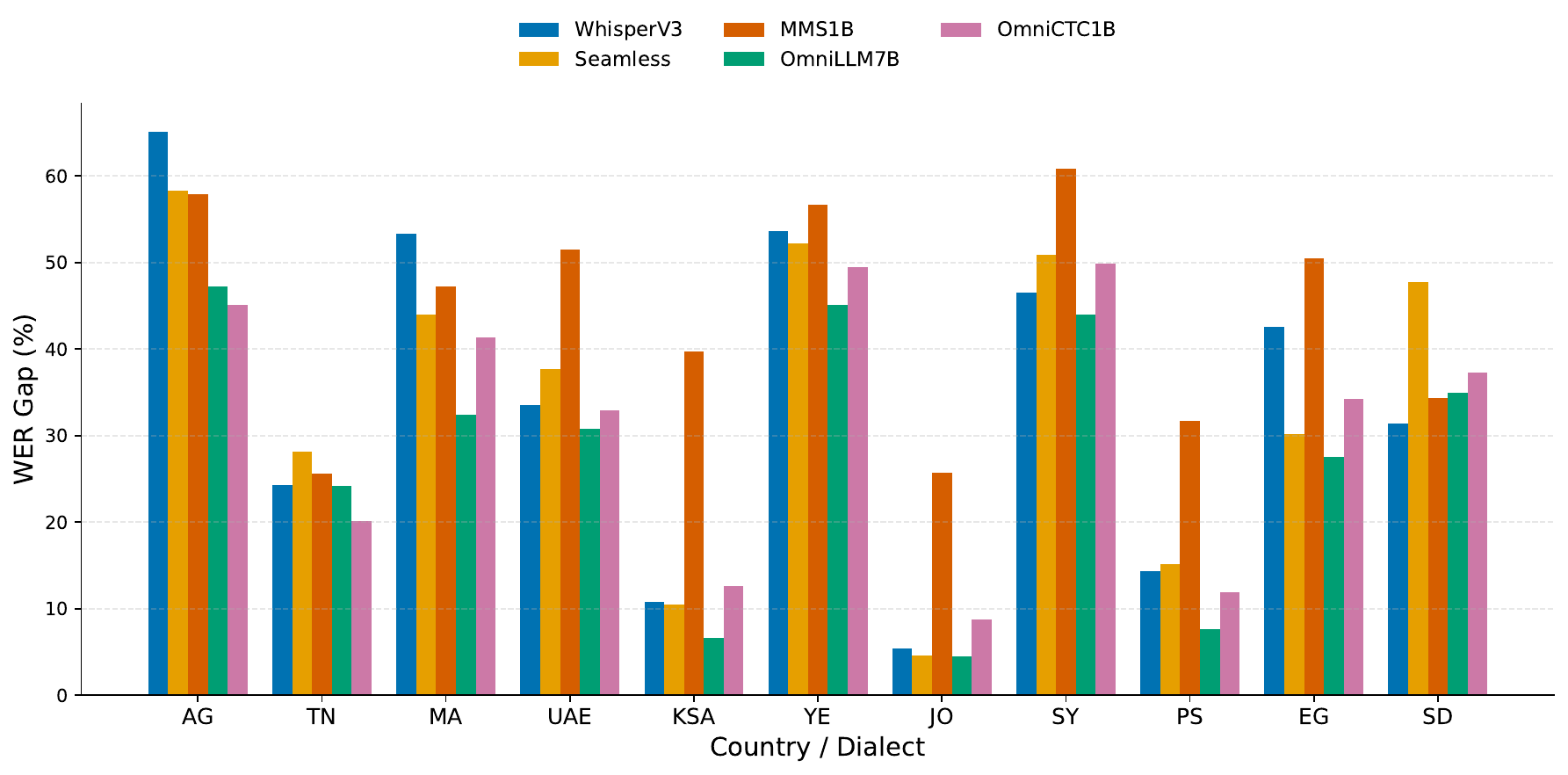}
    \caption{Country-level gap between dialectal Arabic and accented MSA/CA speech. The gap is computed as dialectal WER minus accented WER; larger values indicate weaker cross-variety robustness and greater sensitivity to dialectal variation.}
    \label{fig:gap}
\end{figure}

\vspace{1mm}
\noindent
\textbf{Accented MSA and CA.}
Table \ref{tab:msa_wer_cer} and Figure~\ref{fig:gap} show that ASR performance improves substantially on accented MSA/CA compared to dialectal speech, suggesting that the formal linguistic structure of MSA/CA is generally better aligned with the models’ training distributions, even when spoken with regional accents. \textit{OmniLLM-7B} achieves the best overall performance, with a mean WER/CER of 13.3/4.8, followed closely by \textit{OmniLLM-3B} (13.7/5.1) and \textit{OmniLLM-1B} (14.2/5.3), indicating strong robustness across accent variations. Among general multilingual foundation models, \textit{Seamless} remains competitive (14.4/5.4), while \textit{MMS1B} performs worst (33.8/9.7). Performance still varies by accent. Levantine-accented speech yields the lowest error rates overall, particularly for Syrian-accented speech, where \textit{OmniLLM-7B} achieves 8.1/1.7 WER/CER. In contrast, Sudanese-accented speech remains consistently challenging across all models, with even the strongest systems exceeding 20\% WER. Figure~\ref{fig:gap} further shows that the robustness gap between dialectal and accented speech is largest for North African and Yemeni dialects, indicating that spontaneous dialectal variation introduces substantially greater difficulty than accented formal speech.

\vspace{1mm}
\noindent
\textbf{Model Scale.}
Within the OmniASR-LLM family, performance improves almost monotonically with increasing model size, with larger models consistently yielding lower error rates in both metrics. This indicates that increased model capacity enhances robustness to phonetic, lexical, and acoustic variation.
In contrast, the scaling behavior within the OmniASR-CTC family is less stable. Increasing parameter size does not yield systematic improvements; in some cases, performance degrades slightly. Notably, omniASR-CTC-1B outperforms the larger 3B and 7B variants, indicating that scaling alone does not guarantee gains under the CTC-based architecture. This implies optimization challenges when applying CTC-based decoding at higher parameter counts.

\subsection{Error Analysis}
We observe recurring error patterns including phonetically similar substitutions, dialect-driven phonological mismatches, and distortions of rare or dialect-specific lexical items. Character-level errors often reflect systematic phoneme-to-grapheme mismatches, whereas word-level errors are more common for proper nouns and low-frequency vocabulary. To better understand these failure modes, we conduct a qualitative analysis of the best-performing model (\textit{OmniASR-LLM-7B}) on the test split, revealing three dominant sources of error.

\vspace{1mm}
\noindent
\textbf{Phonetically Similar Substitutions.} The model frequently confuses acoustically similar phonemes, particularly when dialectal realizations diverge from standard pronunciations. For example, emphatic \textipa{/s\textsubwedge{}/} (\armini{ص}) is confused with plain \textipa{/s/} (\armini{س}), producing \armini{صرت} instead of \armini{سرت}. Likewise, pharyngeal \textipa{/Q/} (\armini{ع}) may weaken in connected speech, leading to \armini{ع}$\leftrightarrow$\armini{ا} substitutions. In dialects where \textipa{/q/} (\armini{ق}) is realized as a glottal stop \textipa{/P/}, the model often alternates between \armini{ق} and \armini{ا}, as in \armini{قديش}$\leftrightarrow$\armini{أديش}.

\vspace{1mm}
\noindent
\textbf{Dialect-driven Phonological Variation.} Errors also arise when dialect-specific pronunciation patterns do not align with expected orthographic forms. For instance, the imperfective prefix may appear as \textipa{/b-/} or \textipa{/bi-/} (\armini{بضل} vs.\ \armini{بيضل}), leading to inconsistent transcription. Reduced vowels in fluent speech can cause deletions, such as transcribing \armini{تبارك} as \armini{تبرك}. Similarly, variable realizations of \armini{ج}, such as \textipa{/j/} or \textipa{/tS/}, result in forms like \armini{باتشر} instead of \armini{باجر}.

\section{Conclusions}
\label{sec_conclusions}
In this work, we introduce \ourdataset, a large-scale, community-driven Arabic speech corpus designed to better reflect the linguistic diversity of the Arab world beyond dominant dialects and MSA. \ourdataset provides a challenging and realistic benchmark for Arabic speech technologies under diverse regional and demographic conditions by covering~11 countries and fine-grained sub-dialect annotations, resulting in 16 dialects. Moreover, \ourdataset includes accented MSA and CA recordings, enabling analysis of accent transfer and pronunciation variation across standardized speech forms. Our benchmark evaluation across multiple SOTA ASR systems reveals substantial performance disparities across countries and dialects, highlighting that current speech models remain uneven in their support for underrepresented Arabic varieties. The lexical overlap and speaking-rate analyses further demonstrate the significant heterogeneity across Arabic speech communities, emphasizing the limitations of evaluating models solely on coarse dialect categories. By releasing \ourdataset, we aim to support more inclusive Arabic speech research, enabling future work in dialect-aware ASR, dialect identification, accent adaptation, and speech generation for low-resource Arabic varieties.

\section{Limitations}

\textbf{Emotionally Neutral Recordings}. Due to the absence of surrounding paragraph-level context, participants delivered the sentences in a formal, neutral tone, as the intended emotional expression could not be clearly determined. In addition, the participants were requested to avoid environments with excessive background noise or overlapping speech from other speakers. Consequently, the \ourdataset dataset is limited in its applicability to emotion recognition tasks or speech processing scenarios involving noisy and unconstrained environments. Since the collected data were controlled for emotional expression and background noise, the evaluation of the proposed models is more closely associated with dialectical understanding than with speaker emotion or robustness to environmental noise.

\noindent
\textbf{Country and Gender Imbalance.} Some countries are underrepresented in the dataset, and the gender distribution across dialects is not perfectly balanced. This imbalance is primarily due to the voluntary nature of the data collection process. While this limitation cannot be fully addressed retrospectively, a detailed statistical analysis of the dataset is provided to characterize these imbalances and support a more informed interpretation of the experimental findings.

\noindent
\textbf{Sub-Dialect Coverage Limitation. }Another limitation is the restricted coverage of sub-dialects. The Arab region is characterized by substantial dialectal variation, not only between countries but also within the same country. Comprehensive coverage of all sub-dialects would require collecting speech samples from rural and hard-to-access areas, which presents significant logistical challenges.

\noindent
\textbf{Domain Coverage Imbalance.} Some country subsets exhibit limited domain diversity due to the nature of the data collection. For example, the Palestinian subset is primarily concentrated in the economy domain, resulting in a narrower lexical and topical distribution compared to other country subsets. This limited variation may make the corresponding test data more homogeneous and potentially easier for ASR models, which should be considered when interpreting cross-country performance differences.

\noindent
\textbf{Imbalance in CA and MSA Coverage.} MSA and CA subsets are comparatively limited in size, with recordings contributed by only a small number of participants. This results in both speaker imbalance and reduced diversity in speaking styles, accents, and acoustic conditions. Consequently, findings related to CA and MSA should be interpreted with caution, as the observed performance may not fully generalize to broader speaker populations.

\section{Ethics and Data Statement}
The speech corpus used in this work was collected from speakers across 11 Arab countries to improve dialectal diversity in ASR. All participants provided informed consent for the use of their recordings and transcripts for research purposes. The dataset includes both prompted speech and naturally occurring conversational content, with a subset derived from private chat conversations (WhatsApp and Telegram) voluntarily contributed by their owners. Only conversations for which explicit permission was obtained were included in the dataset.

To protect participant privacy, all conversational texts were carefully anonymized before recording and release. Personally identifiable information (e.g., names, phone numbers, addresses, usernames, email addresses, organizations, and other sensitive references) was removed. Only non-identifying demographic metadata, specifically age and gender, are retained to support research on speaker diversity. No original private chat logs are distributed. Only the anonymized speech recordings and corresponding anonymized transcripts are released.

The dataset is released under a research-only license for non-commercial academic use upon contact with the corresponding author. Users are expected to comply with applicable privacy regulations and ethical guidelines, and the dataset must not be used to identify individuals, reconstruct personal information, or support surveillance or other harmful applications.

Although the dataset covers multiple Arabic dialects and accents, it may still contain demographic and regional imbalances that could affect model performance across different speaker groups. We therefore encourage future work on fairness evaluation and dialect-aware benchmarking. The dataset is intended solely for academic and research use, and we discourage applications that may compromise user privacy or enable harmful surveillance. Overall, this work aims to support more inclusive and representative Arabic ASR systems while adhering to standard ethical practices commonly adopted in NLP and speech research.


\bibliographystyle{acl_natbib}
\bibliography{ref}


\clearpage
\appendix

\section*{Appendix}
\label{sec:appendix}


This appendix provides supplementary material to support the main findings of this work. It is organized as follows:

\begin{itemize}
    \item \textbf{Appendix \ref{appendix:related}: Related Work} \\
    Additional background on multilingual, uni-dialect, and multi-dialect speech datasets.

    \item \textbf{Appendix \ref{appen_transcription_content}: Transcription Content} \\
    Further details of the textual samples recorded in the \ourdataset~dataset. 

    \item \textbf{Appendix \ref{recording}: Speech Recording} \\
    Details of the recording tools and guidelines used to collect the \ourdataset~dataset.

    \item \textbf{Appendix \ref{verification}: Speech Verification} \\
    Detailed explanation of the speech verification tools and guidelines used to verify the recorded speech data.

       \item \textbf{Appendix \ref{data_split}: Dataset Split}  \\
    More information on the \ourdataset~dataset split.

    \item \textbf{Appendix \ref{setup}: Benchmarking Experiment Setup} \\
    Detailed explanation of model configurations, training settings, computational resources, and reproducibility details.

\end{itemize}

\section{Related Work}
\label{appendix:related}
Despite significant advances in Arabic ASR modeling techniques, the availability of diverse, high-quality annotated speech resources remains limited \cite{10.1016/j.specom.2024.103110}. Arabic presents unique challenges due to diglossia, extensive dialectal variation, and frequent code-switching with foreign languages. Moreover, existing datasets differ substantially in linguistic coverage, domain focus, annotation standards, and recording conditions, resulting in fragmented resource landscapes. To contextualize these limitations, we review prior Arabic speech across multilingual, uni-dialect, and multi-dialect datasets.

\paragraph{Multilingual Speech Datasets. }
Multilingual datasets containing Arabic are essential for language identification, mixed-language decoding, and cross-lingual transfer learning. However, relatively few multilingual datasets incorporate Arabic. The GlobalPhone project \cite{schultz02_icslp} introduced a multilingual speech and text dataset encompassing Arabic, collected from Tunisian speakers through prompted newspaper readings. While the dataset offers clean transcriptions and controlled recordings, it is limited to reading text rather than spontaneous conversational data. Similarly, Mozilla Common Voice \cite{ardila-etal-2020-common} is a crowdsourced multilingual dataset containing approximately 15 hours of Arabic speech with demographic metadata including age, gender, and accent, supporting fairness-aware modeling. However, its dialect distribution depended on volunteer participation and therefore lacked systematic balance. In general, while multilingual datasets address cross-lingual modeling challenges, they tend to prioritize language interaction over dialectal diversity within Arabic, limiting their utility for dialect-aware ASR development.

\paragraph{Uni-Dialect Speech Datasets.} 
Uni-dialect datasets target a single regional dialect or accented Arabic variety. Some datasets offer a single Arabic dialect mixed with another language to address the Arabic code-switching problem. For example, FACST corpus \cite{djegdjiga-etal-2018-french} covers Algerian Arabic mixed with French, ArzEn corpus \cite{hamed2021investigationsspeechrecognitionsystems} targets Egyptian Arabic–English code-switching, and TUN-SWITCH \cite{abdallah2023leveragingdatacollectionunsupervised} proposes Tunisian Arabic speech mixed with French and English. 
Beyond code-switching, dialect-specific efforts include the SAAVB dataset \cite{ALGHAMDI200845}, which provides 96 hours of Saudi-accented MSA from 1{,}033 speakers, and the ALGASD dataset \cite{DrouaHamdani2010ALGERIANAS}, which captures regional Algerian diversity from 300 speakers across 11 regions. For Tunisian Arabic, TARIC \cite{Masmoudi2017AutomaticSR}, STAC \cite{zribi-etal-2014-conventional}, and TunSpeech \cite{Messaoudi2021TunisianDE} collectively cover spontaneous, broadcast, parliamentary, and read speech genres. While these Uni-dialect corpora offer valuable depth for dialect-specific modeling, their narrow geographic scope limits generalizability across broader Arabic dialect families. Moreover, most of these datasets are collected from broadcast sources, where the voice is recorded in a controlled environment.  

\paragraph{Multi-Dialect Arabic Speech Datasets.}
Multi-dialect corpora aim to capture variation across multiple Arabic dialects, enabling ASR systems that generalize across regions. One of the earliest efforts, OrienTel \cite{siemund2002orientel}, contains telephone recordings from six Arabic countries with balanced gender representation and diverse acoustic conditions. The Fisher Levantine Arabic corpus \cite{Fesharticle} further contributes approximately 45 hours of spontaneous telephone conversations within the Levantine dialect continuum, covering speakers from Jordan, Lebanon, and Palestine. 

Large-scale broadcast datasets have significantly expanded multi-dialect coverage. The QASR dataset \cite{mubarak-etal-2021-qasr} provides approximately 2{,}000 hours of multi-dialect broadcast speech crawled from Al Jazeera news channel. The MGB challenge series \cite{ ali2017speechrecognitionchallengewild, ali2019mgb2challengearabicmultidialect, Ali_2020} introduced standardized benchmarks drawing from broadcast and online media, with later versions expanding dialect coverage and incorporating web-sourced recordings. Similarly, the MASC \cite{10022652} offers approximately 1{,}000 hours of YouTube-sourced speech, covering multiple dialects and topics, though at the cost of variable recording quality and annotation consistency. The ESCWA.The CS dataset \cite{chowdhury2021modelruleallmultilingual} contains Arabic–English code-switching in a formal setting, drawn from speech collected at official meetings in Algeria, Tunisia, and Morocco, alternating between Arabic and French.

Overall, multi-dialect datasets offer broader linguistic coverage than uni-dialect datasets; however, domain biases toward broadcast speech, uncontrolled acoustic variability in web-sourced data, and inconsistent annotation standards remain significant challenges to robust and generalized Arabic ASR development.

\section{Transcription Content}
\label{appen_transcription_content}
In this section, we present further details of the textual samples recorded in \ourdataset. 

\subsection{Text Pre-processing}
\label{process}
To ensure high-quality and consistent textual data, we developed an automated Arabic text cleaning tool tailored for preprocessing raw corpora. The tool operates on plain text files and applies a multi-stage filtering pipeline. The tool is deployed through an interactive Gradio interface, enabling team members to upload datasets, preview cleaned outputs, and download processed files, thereby streamlining the data preparation workflow.

The preprocessing pipeline starts with Unicode normalization, removal of Arabic diacritics, tatweel, RTL marks, HTML tags, parenthetical text, emojis, and ASCII emoticons, while retaining Arabic letters, relevant punctuation, and Arabic numerals. It then normalizes repeated characters, removes excessive laughter tokens, standardizes whitespace, and trims leading/trailing punctuation. For the Tarjamat MSA/CA subset, additional preprocessing is applied to remove numerical digits, as speakers may verbalize the same number differently, introducing transcription inconsistencies. 

\subsection{Transcription Samples from \ourdataset}
\label{translation}

To illustrate the linguistic diversity captured in \ourdataset, Table~\ref{tab:transcription_samples} presents representative transcription samples from each participating country, along with their English translations.

\begin{table}[t]
\centering
\small
\caption{Representative transcription samples from \ourdataset across participating dialects, with English translations.}
\label{tab:transcription_samples}
\renewcommand{\arraystretch}{1.5}
\begin{tabular}{p{1.0cm} p{5.5cm}}
\toprule
\textbf{Dialect} & \textbf{Text} \\
\midrule

AG &
{\footnotesize \setcode{utf8}\< علابالك بلي فوزي راهوا فاميليا >} \\
& \textit{You know! Fouzi is a family.} \\
\midrule

EG &
{\footnotesize \setcode{utf8}\< يا جماعة! سيبوه ياخذ وقته! >} \\
& \textit{Guys! Let him take his time!} \\
\midrule

JO &
{\footnotesize \setcode{utf8}\< راح استلم راتبي من هاظ الاشي >} \\
& \textit{I will receive my salary from this thing.} \\
\midrule

KSA &
{\footnotesize \setcode{utf8}\< فك الباب ابغى اكلمك شوي >} \\
& \textit{Open the door I want to talk to you a bit.} \\
\midrule

MA &
{\footnotesize \setcode{utf8}\< البطاقة ديالي ما وصلتش لحد دايا >} \\
& \textit{My card hasn’t arrived yet.} \\
\midrule

TN &
{\footnotesize \setcode{utf8}\< ديجا تعدت نص ساعة >} \\
& \textit{It's been more than half an hour.} \\
\midrule

PS &
{\footnotesize \setcode{utf8}\< شو الي غير الي براسك؟ >} \\
& \textit{What changed your mind?} \\
\midrule

SD &
{\footnotesize \setcode{utf8}\<  درب السلامة للحول قريب >} \\
& \textit{Wishing you a safe journy.} \\
\midrule

SY &
{\footnotesize \setcode{utf8}\< شو جاي عبالكم تعملوا؟ >} \\
& \textit{What would you like to do?} \\
\midrule

UAE &
{\footnotesize \setcode{utf8}\< وين سرتوا امس؟ >} \\
& \textit{Where did you go yesterday?} \\
\midrule

YE &
{\footnotesize \setcode{utf8}\< لو سمحت اشتي خدمة الغرف >} \\
& \textit{Excuse me, I want room service.} \\

\bottomrule
\end{tabular}
\end{table}

\subsection{Domain Classification Schema}
\label{domains}
\textbf{Domain Taxonomy Definition.}
To ensure consistent thematic categorization, we defined an 11-category domain taxonomy covering common topical areas observed in the collected Arabic texts: Social, Politics, Economy, Religion, Health, Education, Technology, Entertainment, Sports, Crime, and Other. Each domain was associated with a concise operational definition to guide classification and reduce ambiguity between semantically overlapping categories. The \textit{Other} category was reserved for texts that did not clearly align with any predefined domain and required explicit justification. Table~\ref{tab:domain_taxonomy} presents the full domain definitions used during annotation. The highest number of samples were categorized as social. The large proportion of the social domain is primarily due to the nature of the text collection strategy. A substantial portion of the collected texts consists of everyday conversational messages from communication platforms, such as WhatsApp and Telegram, as well as social media content. We intentionally retained such texts because everyday social communication more naturally reflects dialect-specific vocabulary, expressions, and sentence structures used by native speakers. In contrast, texts from domains such as politics, health, and technology often exhibit a stronger influence from MSA and therefore provide less dialectal variation.

\begin{table*}[t!]
\centering
\small
\caption{Domain taxonomy used for text classification.}
\label{tab:domain_taxonomy}
\begin{tabular}{p{3cm}p{11cm}}
\toprule
\textbf{Domain} & \textbf{Definition} \\
\midrule
Social & Everyday life, family, relationships, gender issues, social norms/traditions, gatherings, interpersonal interactions, and personal opinions about society. \\
Politics & Government, political leaders, public policy, elections, governance, corruption, international relations, and political debate. \\
Economy & Money, employment, inflation, poverty, business, investment, taxes, cost of living, and financial hardship. \\
Religion & Religious teachings, doctrine, worship, religious rulings, theological discussion, and faith-based guidance. Metaphorical mention of ``Allah'' alone is not sufficient. \\
Health & Physical health, illness, mental health, medical advice, hospitals, treatment, and public health. \\
Education & Schools, universities, exams, scholarships, academic life, and learning/teaching processes. \\
Technology & AI, programming, software, platforms/apps, gadgets, cybersecurity, and digital systems. \\
Entertainment & Movies, series, music, celebrities, influencers, and TV/media content. \\
Sports & Teams, matches, competitions, players, and athletic events. \\
Crime & Theft, assault, violence, police/court cases, legal disputes, and criminal incidents. \\
Other & Used only when none of the predefined domains apply; a justification is required. \\
\bottomrule
\end{tabular}
\end{table*}

\begin{figure}[t]
    \centering
    \includegraphics[width=\linewidth]{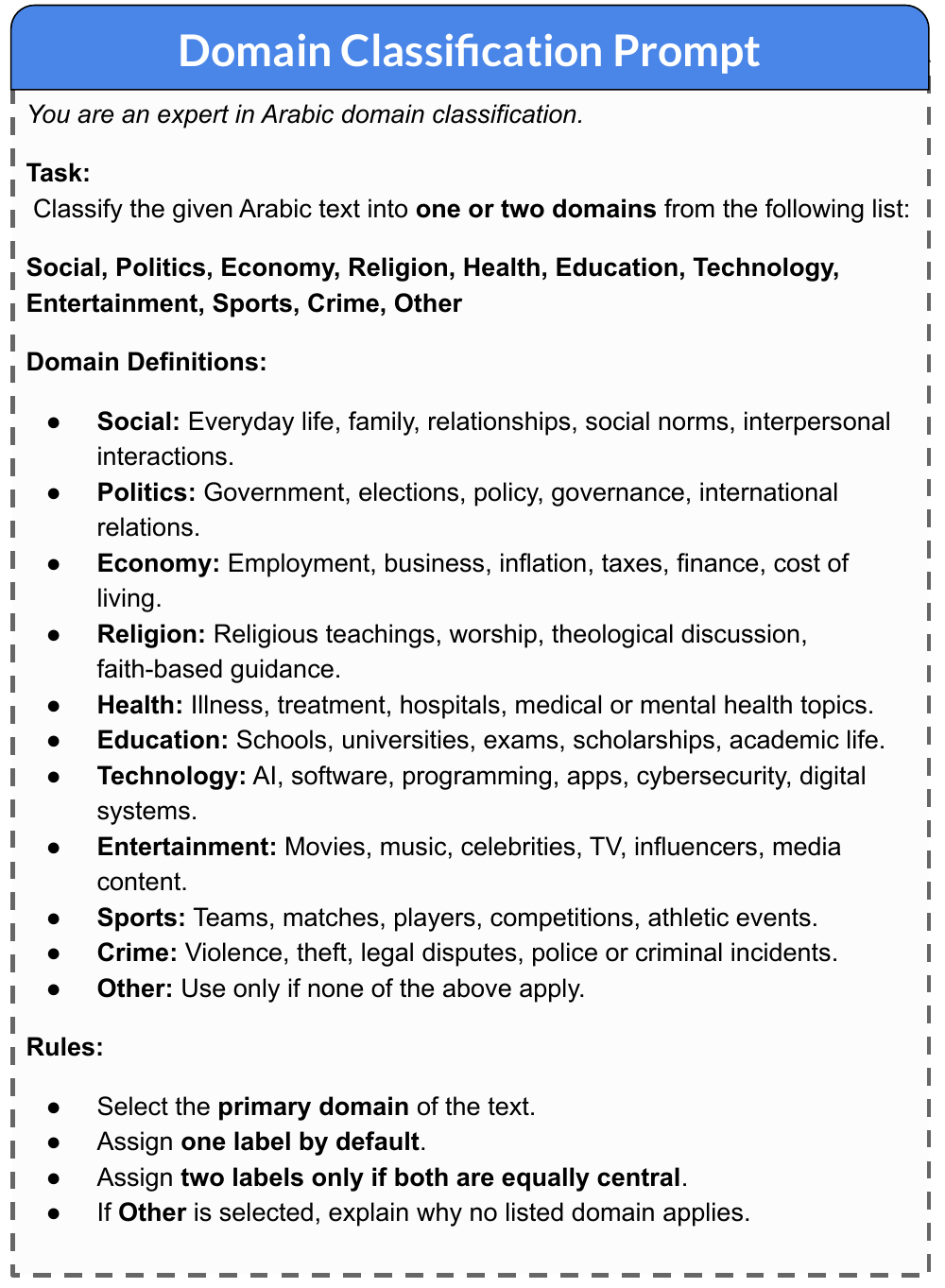}
    \caption{Prompt template used for GPT5-mini domain annotation.}
    \label{fig:domm}
\end{figure}

\noindent
\textbf{Prompt strategy.}
The prompt instructed the model to assign a single primary domain label, allowing up to two labels only when two domains were equally related to the text. If the label \textit{Other} was assigned, the model was required to provide an explicit justification.
The model was also prompted to give a short reasoning statement and a confidence score between 0 and 1, indicating its certainty. Figure \ref{fig:domm} illustrates the prompt used for domain classification.

We manually reviewed and verified all instances with a confidence score below 0.70 to ensure consistent labeling and reduce potential misclassification. The threshold is selected based on an initial manual revision of the classification results.
\\

\begin{figure}[t!]
    \centering
    \includegraphics[width=\linewidth]{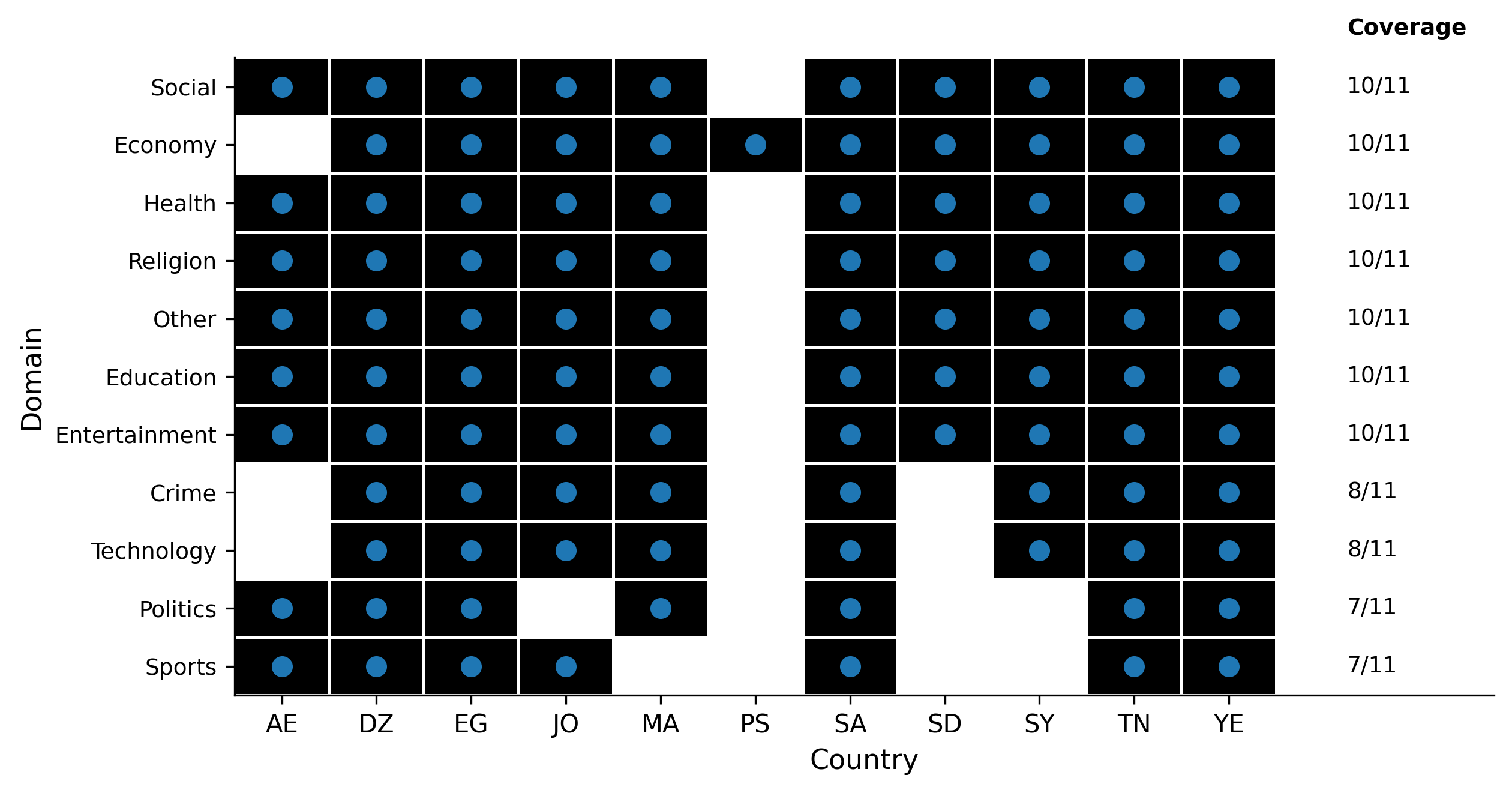}
    \caption{Cross-country domain coverage matrix. Filled markers indicate domain presence, and the rightmost column reports domain coverage across the 11 countries.}
    
    \label{fig:domainava}
\end{figure}

\noindent
\textbf{Domain Distribution Statistics.}
Figure \ref{fig:domainava} further examines domain coverage across countries. Most major domains, including \textit{Social}, \textit{Economy}, \textit{Health}, \textit{Religion}, \textit{Education}, and \textit{Entertainment}, are represented in 10 out of 11 countries, indicating broad topical consistency across regional subsets. More specialized domains, such as \textit{Crime} and \textit{Technology}, appear in 8 countries, while \textit{Politics} and \textit{Sports} are the least represented, each appearing in 7 countries. Palestine differs from other subsets as it consists exclusively of economically oriented utterances. Overall, these findings demonstrate that despite domain imbalance, the dataset achieves broad cross-country topical coverage, supporting robust evaluation under diverse semantic conditions.

\subsection{Lexical and Semantic Analysis}
\label{semantic}
To better capture the linguistic diversity in \ourdataset, we conduct a lexical and semantic analysis across the~11 countries. In addition to cross-country comparisons, we analyze intra-country variation for Saudi Arabia and Yemen, which are further subdivided into sub-dialects.

\begin{figure}[t]
    \centering
    \includegraphics[width=\linewidth]{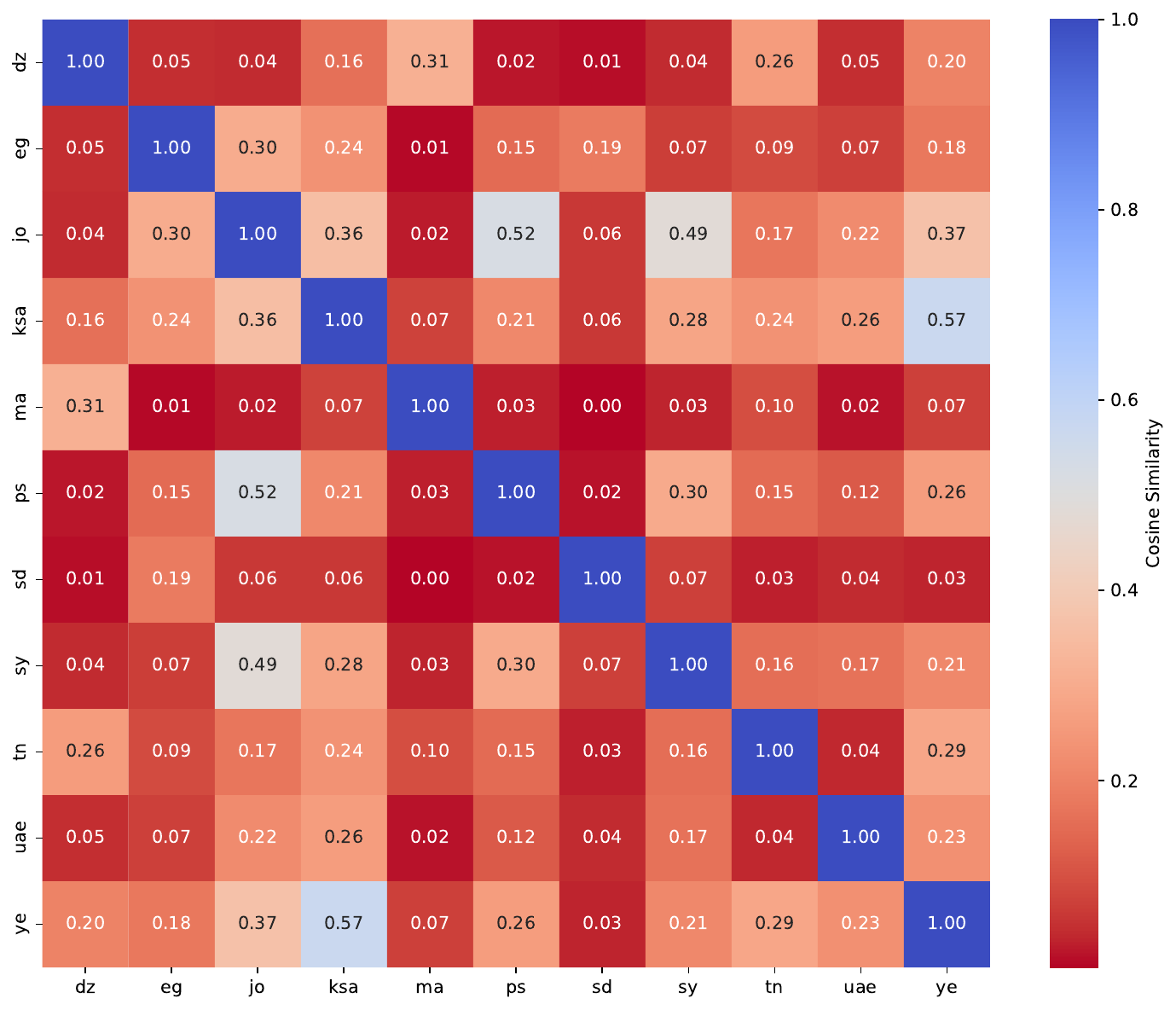}
    \caption{Lexical similarity heatmap across Arabic dialectal corpora in the \textit{BULBUL} dataset, computed using cosine similarity between country-level TF--IDF representations.}
    \label{fig:tfidf}
\end{figure}

\begin{table}[ht]
\centering
\small
\caption{Lexical diversity statistics of the \ourdataset dataset across countries. The table reports the number of sentences (\# Sent.), total number of tokens (Tot. Tokens), vocabulary size (Size; number of unique tokens), type-token ratio (TTR; ratio of unique tokens to total tokens, measuring lexical diversity), and hapax proportion (H Prop.; proportion of tokens that occur only once, indicating lexical richness and sparsity).}
\label{tab:lexical_stats}
\resizebox{\linewidth}{!}{%
\begin{tabular}{lrrrrHHr}
\toprule
\textbf{Country} & \textbf{\# Sent.} & \textbf{Tot. Tokens} & \textbf{Size} & \textbf{TTR} & \textbf{Avg. Word Len.} 
& \textbf{Hapax Count} 
& \textbf{H Prop.} 
\\
\midrule
Algeria   & 632  & 6,055  & 2,811  & 0.464 & 4.21 & 2,108 & 0.750 \\
Egypt     & 1,271 & 13,283 & 4,681  & 0.352 & 3.95 & 3,407 & 0.728 \\
Jordan    & 847  & 8,228  & 3,207  & 0.390 & 3.88 & 2,266 & 0.707 \\
Morocco   & 1,096 & 12,165 & 4,125  & 0.339 & 4.52 & 2,918 & 0.707 \\
Palestine & 1,700 & 14,031 & 2,581  & 0.184 & 4.29 & 1,419 & 0.550 \\
Saudi  & 3,082 & 30,917 & 9,450  & 0.357 & 4.04 & 6,386 & 0.676 \\
Sudan     & 103  & 660    & 499    & 0.756 & 4.13 & 429   & 0.860 \\
Syria     & 1,006 & 6,644  & 3,637  & 0.547 & 4.31 & 2,824 & 0.777 \\
Tunisia   & 429  & 3,503  & 1,949  & 0.556 & 4.04 & 1,551 & 0.796 \\
UAE       & 48   & 306    & 191    & 0.624 & 4.03 & 155   & 0.812 \\
Yemen     & 4,078 & 36,285 & 11,593 & 0.320 & 4.25 & 7,229 & 0.624 \\
\bottomrule
\end{tabular}
}
\end{table}
\noindent
\textbf{Lexical Diversity.}
Table \ref{tab:lexical_stats} presents lexical diversity statistics across countries in the \ourdataset~dataset. Yemen (4,078 sentences; 36,285 tokens) and Saudi (3,082 sentences; 30,917 tokens) sets contain the largest corpora but show moderate type-token ratio (TTR) values of 0.320 and 0.357, respectively. This reflects expected repetition in larger datasets despite their large vocabulary sizes (11,593 and 9,450). In contrast, smaller subsets, such as Sudan (103 sentences) and UAE (48 sentences), exhibit very high TTR values (0.756 and 0.624) and high hapax proportions (0.860 and 0.812), likely inflated by limited data. Countries with broader thematic coverage, such as Syria (TTR = 0.547; hapax prop. = 0.777) and Tunisia (TTR = 0.556; hapax prop. = 0.796), demonstrate strong lexical richness. Notably, Palestine shows the lowest TTR (0.184) and a relatively lower hapax proportion (0.550), as expected, since its text is restricted to a single domain: the Economy. Overall, lexical diversity in the dataset is influenced by both corpus scale and domain heterogeneity. 
\\
\noindent
\textbf{Cross-Country Lexical Overlap.}
Figure \ref{fig:tfidf} presents the cosine similarity heatmap based on TF-IDF representations over the top 100 highest-variance terms. Overall, the heatmap highlights identifiable regional clustering patterns alongside strong cross-dialect distinctiveness across the dataset. The highest similarity is observed between Saudi Arabia and Yemen (0.57), which suggests strong Gulf lexical alignment. Similarly, Jordan shows high similarity with Palestine (0.52) and Syria (0.49), forming a clear Levantine cluster. Some similarity patterns deviate from linguistic expectations due to dataset composition. For example, the low similarity between Syrian and Palestinian dialects is likely due to the domain imbalance.

\begin{figure}[t]
    \centering
    \includegraphics[width=\linewidth]{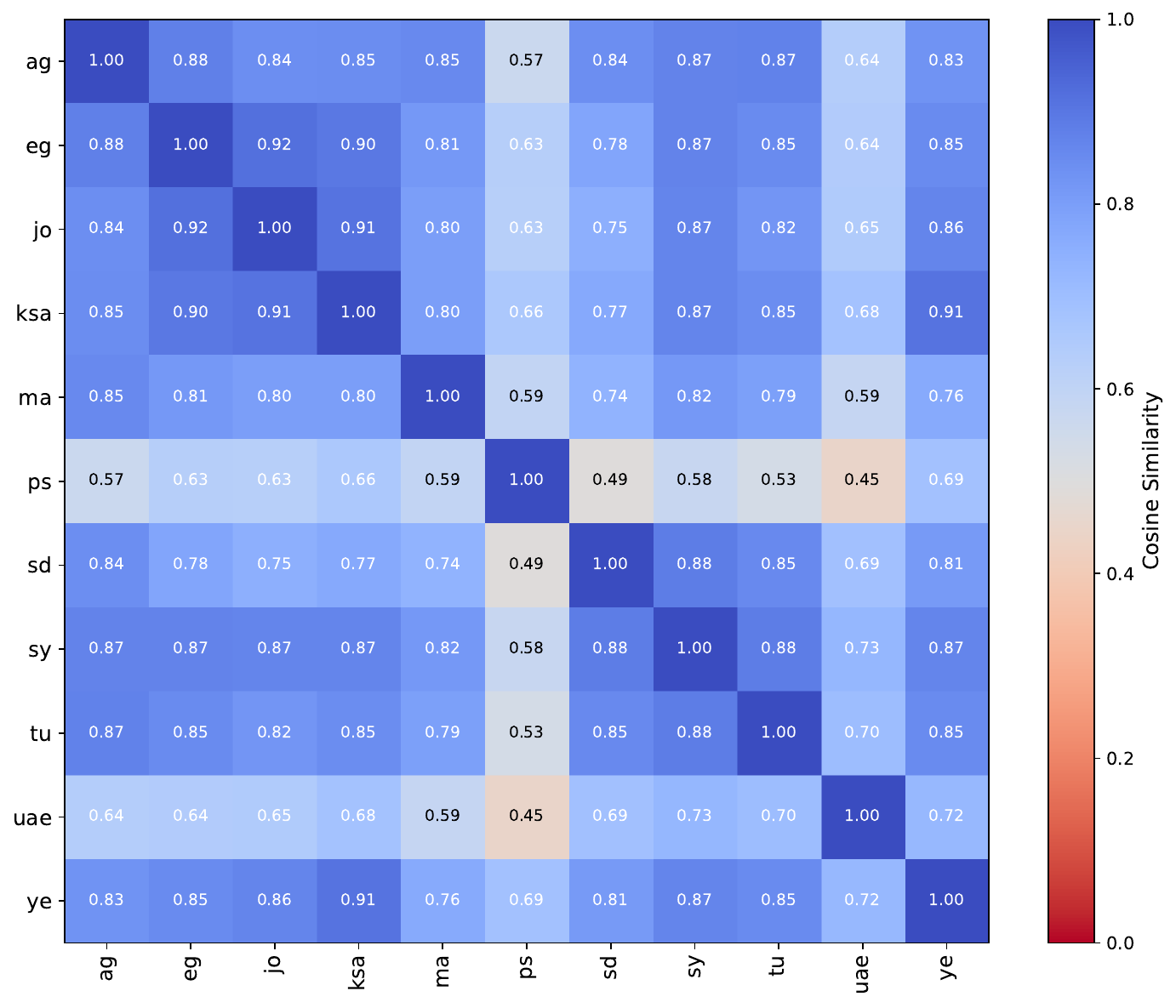}
    \caption{Mean country-level semantic similarity heatmap across the countries from the \textit{BULBUL} dataset, computed using sentence embeddings from the Arabic NLI model \textit{Arabic-arabert-all-nli-triplet}.}
    
    \label{fig:semantic}
\end{figure}

\noindent
\textbf{Semantic Similarity.}
Figure \ref{fig:semantic} illustrates cross-country semantic similarity in the \ourdataset~dataset. We used the Arabic-arabert-all-nli-triplet sentence embedding model to encode each sentence and computed country-level centroids by averaging sentence embeddings. Because the number of available sentences varied substantially across countries,
we limited our computation by randomly sampling \textit{n} sentences per country to avoid corpus-size bias, where \textit{n=48}, which is the minimum number of available sentences across all countries. We repeated the procedure five times with different random seeds and computed cosine similarity between country centroids in each run. Then, we reported the mean similarity matrix across the five runs to ensure robustness.

The results indicate generally high semantic similarity across dialects, mostly greater than 0.75, which reflects a strong shared semantic backbone despite dialectal variation. This can also be attributed to the small subset of samples used in our analysis. A prominent similarity cluster emerges among Jordan, Egypt, Saudi Arabia, Syria, Tunisia, and Yemen. However, Morocco shows slightly lower similarity to the Eastern dialects, consistent with regional divergence. The Palestinian dialect exhibits comparatively lower similarity scores (0.45–0.69), which reflect the topical or lexical characteristics specific to the dataset. Overall, the findings suggest that Arabic dialects, while lexically diverse, remain semantically aligned at the sentence level.

\noindent
\textbf{Sub-dialect Analysis.}
In the Yemeni dialect, we identified 1,464 shared vocabulary items between the Sana'ani and Ta'izzi sub-dialects, including terms such as \textit{"what do you want?"} {\footnotesize \setcode{utf8}\< ماتشتي >}, \textit{"ours"} {\footnotesize \setcode{utf8}\< حقنا >}, and \textit{"leave"} {\footnotesize \setcode{utf8}\< اطرح >}. We also observed systematic lexical differences between the two sub-dialects. For example, Ta'izzi speakers commonly use {\footnotesize \setcode{utf8}\< علشان >}, whereas Sana'ani speakers prefer {\footnotesize \setcode{utf8}\< عشان >} for \textit{"because"}. Similarly, Ta'izzi often uses {\footnotesize \setcode{utf8}\< عليش >} for \textit{"why"}, while Sana'ani typically uses {\footnotesize \setcode{utf8}\< ليش >}.

In the Saudi dialect, many differences between sub-dialects are not lexical substitutions but phonetic or phonological variations. For example, the word \textit{"coffee"} {\footnotesize \setcode{utf8}\< قهوة >} is commonly realized as \textit{gahwa} (\textipa{/gahwa/}) in Najdi speech. In contrast, southern Saudi speakers pronounce it as \textit{ghawa} (\textipa{/ghawa/}), reflecting dialect-specific phonological restructuring that simplifies the consonant sequence. Although both forms correspond to the same orthographic word, they exhibit distinct phonotactic patterns, illustrating how Saudi micro-dialects can diverge in pronunciation even when vocabulary is shared.

\section{Speech Recording}
\label{recording}
This section presents details of the recording tools and guidelines used to record speech data for dialectal and accented formal Arabic. 
\subsection{Recording Tools}
We employed Gradio\footnotetext{\url{https://www.gradio.app/}} to develop a lightweight web-based annotation interface for dataset collection and validation. The interface enabled participants from different countries to record the text corresponding to their country. To ensure high-quality recording, we added the guidelines within the interface. We customized the tool to show each participant's progress and the country's leaderboard to motivate the participants. We also allow participants to review, trim, and repeat the recording before saving it and moving to the next sample. A \textit{Skip} button is also added to enable participants to skip the text and move to another sample.

\subsection{Recording Guidelines}
We provided the participants with guidelines to ensure a controlled environment. The following guidelines were provided to the participants:
\begin{itemize}
    \item \textbf{Environment:} Record in a quiet place and make sure no other human voices appear in the recording. Also, avoid loud background noise that can make the voice inaudible. 
    \item \textbf{Microphone:} It is preferable to use an external microphone or headset, as it is typically clearer than a built-in laptop microphone. If using a mobile phone, ensure the recording quality before proceeding.
    \item \textbf{Speaking Style:} Read each sentence clearly and naturally in your own voice. Do not alter or replace any words unless they reflect natural pronunciation variations (e.g., differences in local pronunciation). If you cannot record a specific sentence or encounter difficulty pronouncing it, you may use the \textit{Skip option} to move to the next sentence.
    \item \textbf{Editing:} You may modify the sentence before starting the recording.
    \item \textbf{Saving:} After completing a recording, click \textit{Save and Next} to save your audio. To re-record, delete the current recording using the \textit{Delete (X)} button. To move to the next sentence without recording, click on \textit{Skip}. 
\end{itemize}
\subsection{Recording Agreement Letter}
All participants who volunteered to join this study were required to agree to the terms outlined in the consent form presented in Figure~\ref{agreement_letter}. The original consent form was presented to participants in Arabic, their native language; the English translation provided here is intended solely for clarity.

\definecolor{paleAgreementBg}{RGB}{255, 242, 242} 
\definecolor{frameBorderColor}{RGB}{220, 200, 200}

\begin{figure*}[t]
\begin{agreementbox}

    \begin{center}
        \large\textbf{Consent Form for Data Collection and Usage}
    \end{center}
    \vspace{-0.5em}
    \setlength{\parskip}{0pt}
    \setlength{\itemsep}{0pt}
    \setlength{\parsep}{0pt}
    {\small

    This agreement is entered into by and between the participant and the research team from King Fahd University of Petroleum and Minerals (KFUPM) and Taibah University (hereinafter referred to as ``the Universities''). The purpose of this agreement is to collect, use, and distribute audio recordings to support audio deepfake detection research and other non-commercial academic pursuits.

    \vspace{0.3em}
    \noindent\textbf{1. Purpose of Data Collection.}
    The research team is collecting audio recordings to construct a dataset dedicated to detecting synthetic voices generated via text-to-speech (TTS), voice conversion (VC), and other generative techniques. These data will be utilized in scientific and academic research to develop advanced methodologies for deepfake detection and related non-commercial research.

    \vspace{0.3em}
    \noindent\textbf{2. Nature of Collected Data.}
    The participant agrees to provide:
    \begin{itemize}
        \item \textbf{Audio Recordings:} Speech samples in their natural voice or generated by reading specified texts/sentences.
        \item \textbf{Optional Metadata:} Demographic details such as gender, age group, dialect, and other relevant metadata.
        \item \textbf{Modification Consent:} Permission to modify, alter, or synthesize their voice using artificial intelligence and speech processing techniques.
    \end{itemize}

    \vspace{0.3em}
    \noindent\textbf{3. Granted Rights.}
    The participant grants the research team the full, royalty-free, and unrestricted right to:
    \begin{itemize}
        \item Record, process, and utilize both their natural voice and any synthetic variants derived from it.
        \item Distribute the resulting dataset (comprising both natural and synthetic audio) to the broader scientific community strictly for non-commercial research purposes.
        \item Publish brief audio samples on professional or academic platforms (such as LinkedIn, X/Twitter, and YouTube) to promote deepfake research awareness or announce dataset availability.
    \end{itemize}

    \vspace{0.3em}
    \noindent\textbf{4. Data Availability and Licensing.}
    The audio dataset (both natural and synthetic components) will be publicly released under a \textbf{Creative Commons Attribution-NonCommercial 4.0 International (CC BY-NC 4.0)} license, allowing researchers to utilize and share the data for non-commercial academic purposes.

    \vspace{0.3em}
    \noindent\textbf{5. Privacy and Confidentiality.}
    \begin{itemize}
        \item The participant's name and any directly identifying personal information will remain strictly confidential and will not be published without explicit written consent.
        \item To ensure anonymity, each participant will be assigned a unique pseudonymized identifier (ID) within the dataset.
    \end{itemize}

    \vspace{0.3em}
    \noindent\textbf{6. Voluntary Participation and Withdrawal.}
    \begin{itemize}
        \item Participation is entirely voluntary (100\%).
        \item Participants retain the right to withdraw from the study or request the deletion of their recordings at any point \textit{prior} to the public release of the dataset.
        \item Once the dataset is publicly released, removal of the data will no longer be possible due to the decentralized nature of open-source data distribution.
    \end{itemize}

    \vspace{0.3em}
    \noindent\textbf{7. Compensation.}
    The participant acknowledges that participation is voluntary and carries no financial compensation. The contribution is made solely to support and advance scientific research.

    \vspace{0.3em}
    \noindent\textit{*By creating an account, you explicitly acknowledge that you have read, understood, and agreed to all the terms and conditions outlined above.}

    }
    \vspace{\fill}
\end{agreementbox}
    \captionof{figure}{Consent form for data collection and usage.}
\label{agreement_letter}
\end{figure*}

\section{Speech Verification}
\label{verification}
This section presents details of the speech verification tools and guidelines used to verify the recorded speech data for dialectal Arabic. 
\subsection{Verification Tool}
To ensure that all recordings adhere to the established guidelines and that the evaluation samples remain well-grounded, we developed a verification tool that enables secondary-speaker oversight during the recording process. The tool allows reviewers to either accept or reject each recording sample. In cases of rejection, the reviewer must select from a predefined set of rejection reasons. This structured process helps ensure that evaluation decisions are justified, consistent, and less susceptible to reviewer bias.

\subsection{Verification Guidelines}
Recordings that did not meet any of these criteria were rejected.
The reviewer could select one or more of the following primary reasons for rejection: \textit{unclear recording}, \textit{text--audio mismatch}, \textit{prolonged silence}, \textit{dialect mismatch}, or \textit{other}. 
If \textit{unclear recording} was selected, the reviewer was additionally required to choose exactly one of the following sub-reasons: the presence of background noise (e.g., music, urban noise, or audible conversations), more than one clearly audible speaker, noticeable echo, or low volume that made the recording difficult to hear. 
Additionally, if \textit{text--audio mismatch} was selected, the reviewer was additionally required to choose exactly one of the following sub-reasons: complete mismatch, partial mismatch, or minor discrepancies (e.g., one or two mispronounced words). Recordings rejected due to \textit{unclear recording} will be published as a separate noise set for researchers interested in evaluating their ASR systems on noisy speech samples.

This tool streamlined the review process by allowing reviewers to systematically evaluate each sample and select from a predefined set of rejection reasons. Such a structured workflow helps reduce bias and improve consistency during evaluation. If none of the predefined reasons adequately describes the issue, reviewers can select \textit{Other} and provide a free-text comment to justify their decision and support their assessment.

\noindent
\section{Dataset Split Statistics}
\label{data_split}
The experimental results presented in Tables \ref{tab:wer_cer} and \ref{tab:msa_wer_cer} report the normalized WER and CER obtained from evaluating different ASR models. In this section, we further describe the distribution and composition of the evaluation samples to provide a more comprehensive understanding of the reported metrics.

Table \ref{tab:split_stats} summarizes the number of utterances, total duration in minutes, and number of speakers for the development and test splits across the evaluated dialect datasets. The development splits were primarily used in the error analysis discussed in Section \ref{sec:analysis}. Table \ref{tab:split_stats} also presents the statistics of the accented MSA/CA dataset. Due to the limited availability of speakers capable of recording accented MSA/CA samples, all recordings for each country were collected from a single speaker.

\begin{table}[t]
\centering
\footnotesize
\renewcommand{\arraystretch}{1.1}
\setlength{\tabcolsep}{3pt}

\caption{Statistics of the dataset splits across dialectal and accented datasets. We report the number of utterances (Utts), total minutes (Mns), and speakers (Spk).}
\label{tab:split_stats}

\resizebox{\linewidth}{!}{%
\begin{tabular}{l|ccc|ccc||ccc}
\toprule
& 
\multicolumn{6}{c||}{\textbf{Dialectal Arabic}} & \multicolumn{3}{c}{\textbf{Accented Arabic}} \\

\multirow{3}{*}{\textbf{dial}} 
& \multicolumn{3}{c|}{\textbf{Dev}} 
& \multicolumn{3}{c||}{\textbf{Test}} & \multicolumn{3}{c}{\textbf{Test}}  \\

\cmidrule(lr){2-4} \cmidrule(lr){5-7} \cmidrule(lr){8-10}

& \textbf{Utts} 
& \textbf{Mns} 
& \textbf{Spk}
& \textbf{Utts} 
& \textbf{Mns} 
& \textbf{Spk}
& \textbf{Utts} 
& \textbf{Mns} 
& \textbf{Spk}

\\

\midrule

Algeria  & 336 & 32.2 & 4 & 303 & 32.3 & 5 & 55 & 17.52 & 1  \\
Egypt  & 385 & 34.3 & 14 & 313 & 34.8 & 16 & 55 & 15.86 & 1\\
Jordan & 607 & 56.9 & 7 & 726 & 57.0 & 7 & 27 & 5.09 & 1 \\
Morocco & 151 & 21.2 & 2 & 201 & 20.4 & 3 & 55 & 17.58 & 1 \\
Palestine & 782 & 69.1 & 4 & 750 & 68.9 & 5 & 55 & 16.13 & 1 \\
Saudi & 121 & 9.6 & 8 & 123 & 9.6 & 8 & 55 & 12.40 & 1 \\
Sudan & 139 & 13.6 & 5 & 162 & 13.2 & 5 & 30 & 10.17 & 1 \\
Syria & 996 & 105.8 & 8 & 1317 & 105.3 & 8 & 55 & 18.11 & 1 \\
Tunisia & 626 & 62.2 & 2 & 744 & 72.2 & 3 & 20 & 5.76 & 1 \\
UAE & 193 & 14.9 & 2 & 538 & 38.0 & 6 & 6 & 1.04 & 1 \\
Yemen & 460 & 50.0 & 17 & 476 & 49.9 & 17 & 55 & 14.98 & 1 \\

\bottomrule
\end{tabular}
}
\end{table}

\section{Benchmarking Experiment Setup}
\label{setup}
Due to computational resource availability, experiments were conducted across multiple computing environments, including Google Colab and institutional server machines equipped with NVIDIA GPUs. The hardware resources utilized in this work included NVIDIA T4, A100, and H100 GPUs on Google Colab, as well as RTX 3090, RTX A6000, and A100 GPUs on servers provided by the SDAIA–KFUPM Joint Research Center. The selection of hardware depended on the experiment scale and resource availability.

All experiments were implemented in Python 3. Since the evaluated systems consisted of three distinct ASR model families with different software requirements and dependencies, separate execution environments were maintained for each model family. On the institutional servers, isolated Anaconda environments were used to manage dependencies and ensure compatibility across models, while separate Google Colab notebooks were configured for corresponding experimental setups.

The experiments in this work focused exclusively on inference using pre-trained ASR models without additional training or fine-tuning. Therefore, variations in hardware configurations primarily affected execution time and resource utilization rather than transcription quality or evaluation outcomes. To ensure fair and reproducible comparisons, all inference experiments were conducted using consistent model weights, inference settings, and software configurations across environments.

\subsection{Text Normalization}
\label{text_norm}

To ensure that the reported CER and WER accurately reflect transcription quality rather than superficial formatting differences, the same text normalization pipeline was applied to both the reference and predicted transcripts. First, Unicode normalization (NFKC) was performed to ensure a consistent character representation. Next, Tatweel (-) characters and Arabic diacritics (Tashkeel) were removed, as they generally do not contribute to the lexical meaning of words in Arabic NLP tasks. Arabic-Indic numerals were then converted to their ASCII equivalents, and all Alef variants (e.g., {\footnotesize \setcode{utf8}\< أ، إ، آ >}) were normalized to the standard Alef ({\footnotesize \setcode{utf8}\< ا >}). Finally, punctuation marks and special symbols were removed, and consecutive whitespace characters were collapsed into a single space after trimming leading and trailing whitespace.

\end{document}